\documentclass{article} % For LaTeX2e
\usepackage{iclr2027_conference,times}
\usepackage[table]{xcolor}
\usepackage{enumitem}
\usepackage[table]{xcolor}
\usepackage{wrapfig}
\usepackage{hyperref}
\usepackage{url}
\usepackage[utf8]{inputenc} % allow utf-8 input
\usepackage[T1]{fontenc}    % use 8-bit T1 fonts
\usepackage{hyperref}       % hyperlinks
\usepackage{url}            % simple URL typesetting
\usepackage{booktabs}       % professional-quality tables
\usepackage{amsfonts}       % blackboard math symbols
\usepackage{nicefrac}       % compact symbols for 1/2, etc.
\usepackage{microtype}      % microtypography
\usepackage{xcolor}         % colors
\usepackage{graphicx}
\usepackage{subfigure}
\usepackage{mathtools}
\usepackage{amsthm}
\usepackage{amsmath}
\usepackage{amssymb}
\usepackage{extarrows}
\usepackage{diagbox}
\usepackage{microtype}
\usepackage{multirow}
\usepackage[most]{tcolorbox}
\usepackage{xcolor}
\usepackage{booktabs}
\usepackage{tabularx}
\definecolor{systemblue}{RGB}{235,243,252}
\definecolor{usergray}{RGB}{245,245,245}
\definecolor{assistantgreen}{RGB}{237,248,240}

\newtcolorbox{systemmessage}{
  colback=systemblue,
  colframe=blue!45,
  boxrule=0.5pt,
  arc=2pt,
  left=3pt,right=3pt,top=2pt,bottom=2pt
}
\newtcolorbox{usermessage}{
  colback=usergray,
  colframe=black!25,
  boxrule=0.5pt,
  arc=2pt,
  left=3pt,right=3pt,top=2pt,bottom=2pt
}
\newtcolorbox{assistantmessage}{
  colback=assistantgreen,
  colframe=green!45!black,
  boxrule=0.5pt,
  arc=2pt,
  left=3pt,right=3pt,top=2pt,bottom=2pt
}
\usepackage{makecell}

\usepackage{amsmath,amsfonts,bm}

\def\eqref#1{equation~\ref{#1}}
\def\1{\bm{1}}
\DeclareMathAlphabet{\mathsfit}{\encodingdefault}{\sfdefault}{m}{sl}
\SetMathAlphabet{\mathsfit}{bold}{\encodingdefault}{\sfdefault}{bx}{n}
\definecolor{text}{rgb}{0.4,0.1,0.4}

\newtheorem{assumption}{Assumption}

\usepackage{algorithm}
\usepackage{algorithmic}

\title{ECG-WM: A Physiology-Informed ECG World Model for Clinical Intervention Simulation}

\author{
Zhikang Chen\textsuperscript{$\dagger$1} \qquad
Yue Wang\textsuperscript{2} \qquad
Sen Cui\textsuperscript{2} \qquad
Yu Zhang\textsuperscript{3} \qquad
Changshui Zhang\textsuperscript{2}
\and
\textbf{Tianling Ren\textsuperscript{2}} \qquad
\textbf{Tingting Zhu\textsuperscript{1}}
\\
\textsuperscript{1} University of Oxford \quad
\textsuperscript{2} Tsinghua University \quad
\textsuperscript{3} Southern University of Science and Technology
\\[2pt]
\textsuperscript{$\dagger$}Corresponding author.
Email: zhikang.chen@eng.ox.ac.uk
}
\makeatletter
\renewcommand{\headrulewidth}{0pt}  % 去掉页眉的横线
\makeatother
\iclrfinalcopy % Uncomment for camera-ready version, but NOT for submission.
\begin{document}

\maketitle

\begin{abstract}
Electrocardiogram (ECG)-based models have achieved strong diagnostic
performance, yet remain limited in modeling how cardiac dynamics evolve
under external interventions. Existing approaches primarily focus on
static prediction, with limited capacity to capture ECG changes under
different pharmacological conditions. In this work, we propose an ECG
World Model  for action-conditioned predictive simulation of cardiac
electrophysiology. Our framework integrates physiological ordinary
differential equation (ODE) priors into latent diffusion dynamics through
energy regularization. This structural guidance encourages the generation
of physiologically plausible post-intervention ECG trajectories while
reducing generative inconsistencies. Building on these simulations, we
introduce an uncertainty-aware evaluation strategy that leverages
stochastic diffusion sampling to estimate both the expected risk score
and its variability, supporting comparative assessment of candidate
interventions. We evaluate our method across diverse settings, including
controlled drug-response datasets and real-world clinical records.
Beyond improvements in waveform prediction, experimental results
demonstrate better preservation of clinically relevant ECG intervals
and more accurate prediction of the magnitude and direction of
drug-induced risk changes. These results highlight the potential of
our approach for intervention-aware cardiac simulation and
simulation-based assessment of candidate treatments.
\end{abstract}

\section{Introduction}

%%% 目前随着心电等技术的不断进步，

% Recent advances in foundation models have revolutionized intelligent systems across modalities \citep{wei2022chain, li2023llava}. However, a critical gap remains: while current models excel at pattern recognition, their capacity for complex clinical decision-making is still in its infancy. In cardiology, Electrocardiography (ECG) is a cornerstone of diagnosis, yet its utility extends far beyond simple arrhythmia classification. Clinicians must synthesize ECG signals with longitudinal patient data to make high-stakes intervention decisions, such as selecting antiarrhythmic therapies or rate-control medications. While emerging ECG foundation models \citep{mckeen2025ecg, xu2026ecg} have achieved state-of-the-art performance in automated detection, they often function as "black-box" classifiers rather than clinical reasoning partners.

%Recent advances in large-scale foundation models have demonstrated remarkable emergent capabilities across a wide range of domains \citep{wei2022chain, li2023llava, gu2024mamba, shao2024deepseekmath}, including natural language understanding, visual perception, and complex reasoning. These models have shown strong generalization across tasks and modalities, enabling new paradigms for intelligent systems. However, despite these advances, a critical gap remains:

Recent advances in large-scale foundation models have demonstrated remarkable emergent capabilities across a wide range of domains \citep{wei2022chain, li2023llava, gu2024mamba, shao2024deepseekmath}. However, extending these capabilities to physiologically grounded simulation in high-stakes, dynamic medical domains remains challenging.

\begin{tcolorbox}[
colback=white,
colframe=gray,
arc=3mm,
boxrule=0.8pt,
shadow={0.5mm}{-0.5mm}{0mm}{black!50!white}
]
\itshape
\textbf{The Clinical Imagination Gap:} While current AI excels at parsing the clinical past, reliable simulation of intervention-conditioned futures remains challenging. Viewing clinical treatment as a Partially Observable Markov Decision Process (POMDP) motivates extending medical AI from static pattern recognition to action-conditioned physiological simulation.
\end{tcolorbox}

\begin{figure}[t!]
    % \vspace{-.1cm}
    % \setlength{\abovecaptionskip}{-10cm}
    % \setlength{\belowcaptionskip}{-10cm}
    \centering
    \includegraphics[width=1\columnwidth]{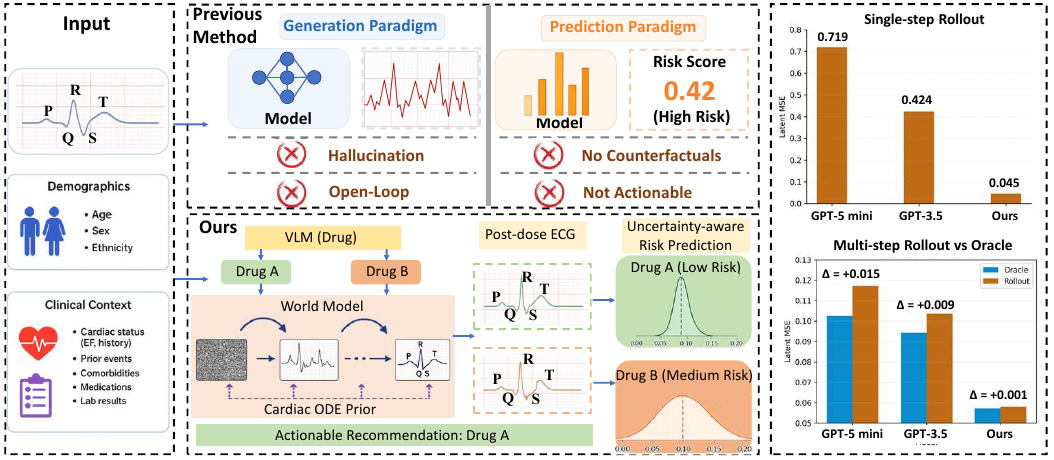}
    \vspace{-0.5cm}
    \caption{Overview and evaluation paradigm. Left: Existing ECG generation and prediction methods primarily operate in open-loop or static settings, whereas our framework models intervention-conditioned post-dose trajectories and evaluates simulated outcomes. Right: Single- and multi-step evaluations quantify predictive fidelity and rollout--oracle terminal-error differences under intervention-conditioned prediction.}
    \label{fig:intro}
    \vspace{-.6cm}
\end{figure}

This limitation is especially pronounced in high-stakes medical domains, where effective clinical analysis requires not merely pattern recognition, but a mechanistic understanding of how physiological processes respond to interventions. Electrocardiography (ECG), a fundamental tool in cardiology, provides a rich representation of cardiac electrical activity and is routinely used to support diagnosis and monitor disease progression. Despite recent progress in ECG foundation models \citep{mckeen2025ecg,xu2026ecg,knight2026wearable}, which achieve strong performance on tasks such as arrhythmia classification and disease prediction, these approaches remain largely focused on discriminative or predictive objectives. They often provide limited support for characterizing how physiological states evolve under clinical interventions, such as medications or therapies. This limitation becomes particularly critical in interventional settings. In practice, clinicians must consider multiple candidate medications, requiring prospective \textit{``what-if'' reasoning} to anticipate how a patient's cardiac state might evolve under each option. Since real-world trial-and-error carries clinical risks and responses to alternative treatments cannot all be observed for the same patient under identical conditions, prospective simulation offers a route to more individualized analysis. Current ECG-based AI systems provide limited support for such prospective reasoning due to their fragmented design. As illustrated in Figure~\ref{fig:intro} (left), generative models focus primarily on open-loop signal reconstruction, while predictive models estimate risk from static observations without explicitly modeling treatment actions. A more comprehensive framework would integrate (i) candidate pharmacological actions, (ii) simulation of action-conditioned cardiac dynamics, and (iii) evaluation of downstream ECG-derived risk signals. Without explicitly integrating these components, current AI systems remain limited in supporting simulation-driven analysis in cardiology.

To address the aforementioned challenges, we introduce an ECG World Model for action-conditioned predictive simulation of cardiac electrophysiology. As illustrated in Figure~\ref{fig:intro} (left), rather than simply pipelining disjoint networks, our framework integrates non-linear physiological ordinary differential equation (ODE) priors into latent diffusion models through EPK conditioning and physiology-informed energy regularization. Instead of directly predicting clinical outcomes from static signals, our approach explicitly models the action-conditioned evolution of cardiac states. Given an input ECG representing the current cardiac state, the system first proposes candidate pharmacological actions. Guided by these ODE priors, our diffusion-based dynamics model is trained to encourage physiologically plausible post-intervention ECG states. Subsequently, these simulated trajectories are evaluated using a frozen ECG foundation model to obtain diagnostic scores that serve as risk proxies. Crucially, to bridge the gap between generative simulation and downstream analysis, we introduce an uncertainty-aware evaluation strategy. Instead of relying on a single deterministic estimate, our framework leverages the stochasticity of the diffusion process to sample multiple potential future trajectories. This enables the characterization of both the mean diagnostic risk score and its variability across simulated trajectories for each candidate intervention. As a result, our approach extends ECG analysis from a purely diagnostic setting to a simulation-driven framework for comparative evaluation. Furthermore, the rollout--oracle comparison in Figure~\ref{fig:intro} (right) shows small additional average terminal errors over the evaluated short chains. Complementary experiments demonstrate improved waveform and interval fidelity and stronger agreement with observed post-dose diagnostic changes, supporting the potential of this framework for intervention-aware cardiac analysis. The main contributions of this paper are summarized as follows:

\begin{itemize}[leftmargin=10pt]
\vspace{-6pt}
    \item[$\bullet$] \textbf{Physiology-Informed World Model}
    We propose an ECG World Model that integrates non-linear cardiac ODE priors into a latent diffusion process via energy regularization, encouraging physiological consistency in generated trajectories. Our model combines structured physiological priors with learned multi-lead waveform representations.

    \item[$\bullet$] \textbf{Intervention-Conditioned Simulation}
    We unify candidate treatment generation, post-intervention ECG simulation, and downstream risk-proxy evaluation into an integrated framework for comparative intervention analysis. This framework enables intervention-aware analysis by modeling the potential physiological responses to different medications.

    \item[$\bullet$] \textbf{Uncertainty-Aware Risk Evaluation}
    By leveraging the stochasticity of the diffusion process, our framework characterizes the mean and variability of diagnostic risk proxies for candidate interventions. These quantities support simulation-based candidate comparison, while diagnostic-change fidelity is assessed against observed post-dose ECGs.

    \item[$\bullet$] \textbf{Comprehensive Evaluation}
    Our method improves waveform prediction, ECG interval preservation, and diagnostic-change agreement over the evaluated baselines across datasets covering normal and abnormal cardiac states. Additional analyses assess short-chain recursive prediction and reconstruction under missing-lead conditions. The project page is available at \url{https://chenzk202212.github.io/ECG_World_Model/}.
\end{itemize}

\vspace{-.35cm}
\section{Related Work}
\vspace{-.3cm}

\paragraph{World Models for Decision-Making} 
In recent years, World Models \citep{ha2018world, matsuo2022deep, zhou2025hermes, chen2026generative} have attracted increasing attention for their ability to simulate the dynamic evolution of environments. Recent advances incorporate explicit causal structures, physiological constraints, and high-dimensional control. For example, SPARTAN \citep{leispartan} enforces causal stability via time-dependent causal graphs, while PIN-WM \citep{li2025pin} embeds differentiable rigid-body dynamics to ensure physical consistency. In terms of controllability, Ctrl-world \citep{guo2025ctrl} improves action-conditioned generation, and 3D-VLA \citep{zhen20243d} integrates 3D perception with embodied actions. Beyond open-loop simulation, the Dreamer family \citep{embodiedreamer} enables closed-loop learning of latent dynamics and policy optimization through imagination. Such capabilities are particularly relevant in healthcare settings, where trial-and-error is costly and often irreversible, making the modeling of intervention effects an “ultimate epistemic stress test” for world models \citep{chen2026generative}. Early explorations in this domain have recently emerged: VCWorld \citep{wei2025vcworld} constructs a white-box biological simulator for gene-level interventions, while MeWM \citep{mewm} applies world models to clinical imaging by generating treatment hypotheses, simulating tumor progression, and evaluating outcomes via survival analysis. However, despite these advances in imaging and genomics, the extension of world models to continuous physiological signals (e.g., ECG) for intervention-aware simulation and analysis remains largely unexplored. To address this gap, we propose an ECG World Model for modeling action-conditioned cardiac dynamics.

\vspace{-.3cm}
\paragraph{Static and Conditional ECG Generation}  
Early ECG research focused mainly on signal synthesis and risk prediction \citep{chen2024novel, lai2025diffusets, adib2025synthetic}, such as diffusion-based data augmentation \citep{adib2023synthetic}, multi-lead reconstruction \citep{liu2024synthesis}, and cardiovascular risk prediction \citep{zhou2025hybrid, prifti2021deep}. More recent work introduces conditional and multimodal generation: BeatDiff \citep{bedin2024leveraging} formulates ECG reconstruction as a Bayesian inverse problem, and DiffuSETS \citep{lai2025diffusets} incorporates clinical text and metadata for personalized ECG synthesis. %To model intervention effects, Cardiac Digital Twins \citep{camps2025harnessing} combine MRI and ECG with mechanistic models, while DADM \citep{shao2025generation} integrates ODE-based physiological priors to simulate drug-induced ECG changes. However, DADM remains a passive, open-loop signal generator. It does not evaluate the clinical consequences of the waveforms it synthesizes, nor can it autonomously compare alternative treatments. We bridge this gap by introducing an ECG World Model that transitions from passive synthesis to a closed-loop decision-making framework. Given a patient's baseline state, our model actively proposes candidate interventions, simulates their physiological futures, and evaluates them under uncertainty to rank the safest clinical action. 
To model interventions, Cardiac Digital Twins \citep{camps2025harnessing} and DADM \citep{shao2025generation} leverage mechanistic and ODE-based priors. However, these remain passive, open-loop generators that cannot evaluate clinical outcomes or compare treatments. We bridge this gap with a closed-loop ECG World Model. Given a baseline state, our framework actively proposes candidate interventions, simulates their physiological trajectories, and evaluates their simulated outcomes to compare
diagnostic-proxy scores under sampling variability.

\vspace{-.3cm}
\section{Preliminaries}
\vspace{-.2cm}
\label{sec:3}
\subsection{ECG-based Treatment Decision Problem}

Electrocardiography (ECG) signals provide a non-invasive measurement of cardiac electrophysiological activity and play an important role in analyzing responses to pharmacological interventions. Given an ECG observation, the goal is to characterize how candidate medications may influence patient-specific cardiac dynamics under different intervention scenarios. 
Formally, an ECG recording is represented as $x \in \mathbb{R}^{C \times T_{\text{sig}}}$, where $C$ denotes the number of channels and $T_{\text{sig}}$ denotes the temporal length of the signal. An encoder $f_{\text{enc}}(\cdot)$ maps the ECG signal to a representation in the latent space: $z = f_{\text{enc}}(x)$, where $z \in \mathbb{R}^{d}$ captures the underlying physiological state of the patient. 
Let $\mathcal{A}$ denote the set of candidate pharmacological interventions, where each action $a \in \mathcal{A}$ corresponds to a single medication or a multi-step therapy. Deterministic approaches typically evaluate interventions using a static scoring function $R(z,a)$, where $R(\cdot)$ denotes a clinical risk estimate associated with a simulated outcome. Rather than directly optimizing for a single intervention, our focus is on modeling and comparing the potential outcomes induced by different candidate interventions. As we demonstrate, deterministic point-estimates of interventions are fragile in safety-critical clinical settings, motivating an uncertainty-aware formulation that captures both the expected risk and the variance across simulated trajectories.

% \vspace{-.3cm}
\subsection{Denoising Diffusion Probabilistic Models and Energy Regularization}

Denoising Diffusion Probabilistic Models (DDPMs) \citep{ho2020denoising} offer a powerful paradigm for generative modeling by learning to invert a predefined Markovian noise-addition process. Let $z_0$ be the initial clean latent representation. The forward process gradually adds Gaussian noise over $N$ steps, producing a sequence of increasingly noisy states $z_1, \dots, z_N$. The generative reverse process then learns the transition probability $p_\theta(z_{\tau-1} | z_\tau, a)$ to iteratively denoise a Gaussian prior $z_N \sim \mathcal{N}(0, I)$ back to the target distribution, conditioned on an external variable such as an action $a$, where $\tau \in \{1, \dots, N\}$ denotes the internal diffusion timestep. 

To incorporate structural constraints or external physical knowledge into DDPMs without retraining the core noise-prediction network, \textit{energy regularization} is a mathematically principled technique. Given a condition (e.g., an action $a$) and an energy function $\mathcal{E}(z_\tau, a)$ that measures the deviation of the intermediate state from the desired constraint, the generative reverse process can be steered using the gradient of this energy. The modified score function is formulated as:
\begin{equation}
\nabla_{z_\tau} \log \tilde{p}(z_\tau | a) = \nabla_{z_\tau} \log p_\theta(z_\tau | a) - \gamma \nabla_{z_\tau} \mathcal{E}(z_\tau, a),
\label{eq:energy_reg_prelim}
\end{equation}
where $\gamma > 0$ is a temperature coefficient, which controls the guidance scale. This formulation effectively tilts the original data distribution toward regions that minimize the energy function, allowing for flexible constraint-satisfaction during the step-by-step stochastic generation. 
Energy-based score modification provides a useful probabilistic interpretation of how physiological constraints can bias diffusion trajectories. In our implementation, however, we do not apply this gradient correction explicitly during inference. Instead, the physiological energy is incorporated as a training-time regularizer on the predicted clean latent state. Consequently, the deployed model uses standard DDPM sampling, with physiological information amortized into the learned denoising dynamics.

\subsection{Mechanistic Modeling of Cardiac Electrophysiology}

Although deep learning captures statistical patterns, classical electrophysiology provides mechanistic priors. A foundational continuous-time model for generating synthetic ECGs is the McSharry dynamical system \citep{mcsharry2003dynamical}. It characterizes the intrinsic dynamical properties of the ECG signals using an Ordinary Differential Equation (ODE). 
The continuous-time evolution of the theoretical ECG signal amplitude $y(\theta)$ along the cardiac phase $\theta$ is formulated as:
\begin{equation}
\frac{dy(\theta)}{d\theta} = - \sum_{i \in \{P,Q,R,S,T\}} \alpha_i \Delta \theta_i \exp\left(-\frac{\Delta \theta_i^2}{2b_i^2}\right) - (y(\theta) - y_0),
\label{eq:ode_prior_prelim}
\end{equation}
where $\Delta \theta_i = (\theta - \theta_i) \bmod 2\pi$ defines the phase shift corresponding to specific anatomical landmarks (P, Q, R, S, T waves), $\alpha_i$ and $b_i$ control the amplitude and width of these waves, and $y_0$ represents the baseline wander. Let $\theta \in [0, 2\pi)$ denote the continuous cardiac phase. This phase-space ODE provides a robust structural template for both healthy rhythms and pathological variations. 
It is important to note that $y(\theta)$ denotes a $1$-dimensional theoretical scalar amplitude (such as Lead II). We explicitly use $y$ to distinguish this abstract, single-channel physical template from the raw multi-lead observation matrix $x \in \mathbb{R}^{C \times T_{\text{sig}}}$. Rather than explicitly formulating the complex 3D spatial projections of all 12 leads within the ODE, we leverage $y(\theta)$ purely as a \textit{global temporal rhythm template}. This 1D template captures fundamental mechanistic dynamics, which are subsequently mapped into the latent space by our networks to guide the high-dimensional multi-lead generation.

\subsection{Risk-Sensitive Evaluation under Uncertainty}

In standard predictive tasks, actions are often evaluated based on a single point-estimate of the expected outcome. However, in safety-critical domains, such as medical interventions, the inherent stochasticity of patient responses requires an uncertainty-aware evaluation. 
Following standard risk-sensitive decision theory, let $R(z, a)$ be a risk evaluation function. For a stochastic predictive model that produces a distribution of potential future trajectories, $K$ independent samples can be drawn to estimate the empirical mean $\mu_R(a)$ and the standard deviation $\sigma_R(a)$ (derived from variance $\sigma_R^2(a)$) of the clinical risk. To balance the expected risk against its variability, the evaluation is typically formulated using a mean-variance objective:
\begin{equation}
S(a) = \mu_R(a) + \lambda \cdot \sigma_R(a),
\label{eq:final_decision_prelim}
\end{equation}
where $\lambda > 0$ is a risk-aversion hyperparameter. This objective penalizes interventions that have highly variable or unpredictable outcomes, prioritizing those with low expected risk.

\begin{figure}[t!]
    % \vspace{-.1cm}
    % \setlength{\abovecaptionskip}{-10cm}
    % \setlength{\belowcaptionskip}{-10cm}
    \centering
    \includegraphics[width=1\columnwidth]{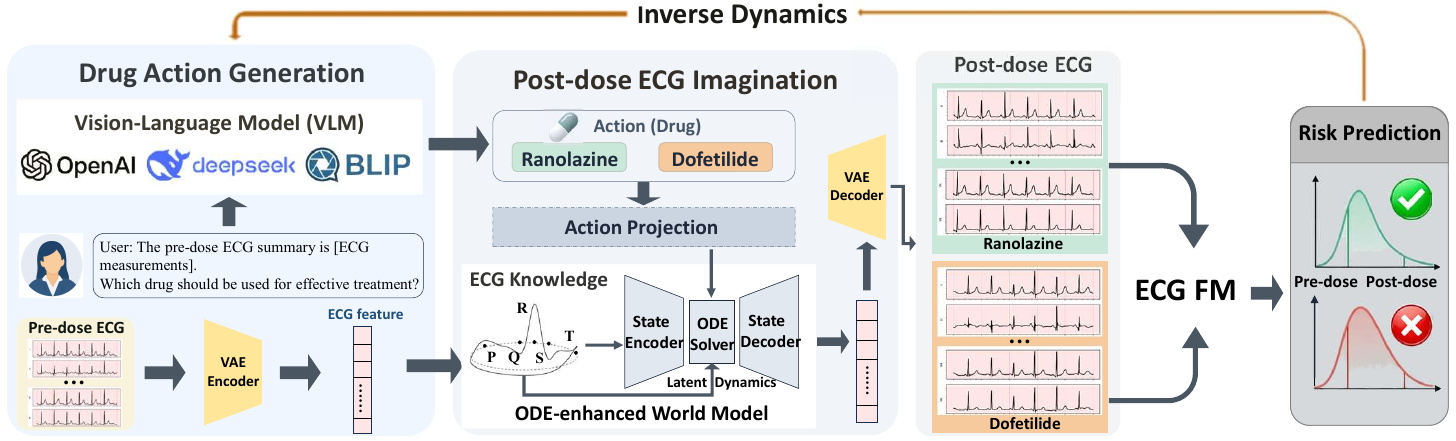}
    \vspace{-0.6cm}
\caption{
Overview of our intervention evaluation framework.
Candidate drugs, optionally proposed by a VLM, are evaluated
through ODE-enhanced post-dose ECG simulation and downstream
diagnostic-proxy scoring.
The reported experiments use fixed candidate sets.
}
    \label{fig:intro1}
    \vspace{-.3cm}
\end{figure}

\section{Methodology}
\label{sec:4}

%Section~\ref{sec:framework} introduces the overall ECG World Model framework, which establishes a simulation-driven pipeline for action-aware cardiac analysis. The framework integrates latent state representation, physiology-guided dynamics modeling, and risk-based evaluation into a unified closed-loop system. Section~\ref{sec:world_model} details the proposed \textit{Physiology-Informed Energy Regularization} mechanism, which imposes differential equation constraints onto diffusion-based latent dynamics, effectively mitigating generative hallucinations. In Section~\ref{sec:action}, we introduce a clinically constrained action generation strategy to ensure pharmacological validity. Finally, Section~\ref{sec:risk} presents the \textit{Uncertainty-Aware Evaluation Module}, which leverages diffusion stochasticity to characterize risk under uncertainty for comparative analysis.

\subsection{The Overall Framework}
\label{sec:framework}

To enable action-aware cardiac analysis, we propose the \textbf{ECG World Model}, an end-to-end framework designed to simulate the evolution of cardiac electrophysiology under pharmacological interventions. As illustrated in Figure \ref{fig:intro1}, our framework extends ECG analysis from purely diagnostic settings to a predictive simulation-based paradigm. Specifically, given a pre-dose ECG signal $x_k$ at clinical step $k$, patient metadata $d$, and a candidate pharmacological intervention $a$, dedicated encoders extract the baseline cardiac state $s_k$, the clinical condition embedding $c_d$, and the intervention embedding $c_a$. Building upon these representations, the core of our world model leverages a hybrid dynamics architecture to simulate the \textit{action-conditioned post-dose ECG} $x_{k+1}$. To mitigate physiological inconsistencies, an \textit{External Physiological Knowledge (EPK)} module first translates the intervention $a$ into a physiology-guided prior $e_{\text{epk}}$ by solving the underlying cardiac ODE. This prior is injected into a Denoising Diffusion Probabilistic Model (DDPM), which iteratively transforms a Gaussian noise distribution into a constraint-aware post-dose ECG sample. Once simulated, the post-intervention physiological trajectory $x_{k+1}$ is evaluated by a downstream \textit{ECG Foundation Model}. By performing multiple stochastic simulations across candidate interventions, the framework enables an uncertainty-aware characterization of both expected risk and its variability, supporting comparative evaluation of different intervention scenarios.

\subsection{Physiology-Informed ECG World Model}
\label{sec:world_model}

Directly modeling high-dimensional ECG signals is challenging due to complex temporal dependencies. We encode raw ECG signals into a low-dimensional latent space using a pre-trained variational autoencoder (VAE). Given an ECG signal $x$, the encoder maps it to a latent variable $z = f_{\mathrm{enc}}(x) \in \mathbb{R}^d$. The decoder reconstructs the signal as $\hat{x} = f_{\mathrm{dec}}(z)$. The VAE is trained by minimizing the standard reconstruction loss and the KL divergence. This compact state space enables efficient modeling of cardiac dynamics under pharmacological interventions.
To incorporate physiological constraints into latent cardiac dynamics and mitigate generative inconsistencies, we propose an energy-regularized diffusion world model. Let $k$ denote the clinical intervention step. An idealized energy-tilted reference distribution is
\begin{equation}
p^\star(z_{k+1} \mid z_k, a)
\propto p_{\mathrm{diff}}(z_{k+1} \mid z_k, a)
\exp\!\big(-\gamma E_\phi(z_{k+1})\big),
\end{equation}
where $p_{\mathrm{diff}}$ is the reference diffusion distribution, $\gamma>0$ controls the theoretical tilt, and $E_\phi$ penalizes deviations from a physiological anchor. This formulation motivates regularized training rather than exactly characterizing the deployed sampler (Appendix~\ref{app:theory_proofs}).

\paragraph{Latent EPK Energy}
Given the 1D continuous physiological trajectory derived from the ODE-based cardiac model, we discretize it into a sequence $e_{\mathrm{EPK}}$. To align this temporal rhythm with the latent space, we apply a projection module $\Pi_\phi(\cdot)$, implemented via temporal resampling and a learnable $1\times1$ convolution, to obtain $z_{\mathrm{EPK}} = \Pi_\phi(e_{\mathrm{EPK}})$. For a fixed clinical transition, let $z_0$ denote the clean target latent and $\tau \in \{1,\dots,N\}$ the internal diffusion timestep. We define the energy on the predicted clean latent $\hat{z}_0 = \frac{z_\tau-\sqrt{1-\bar{\alpha}_\tau}\epsilon_\theta}{\sqrt{\bar{\alpha}_\tau}}$:
$
E_\phi(\hat{z}_0)
=
\left\|\hat{z}_0-\Pi_\phi(e_{\mathrm{EPK}})\right\|_2^2.
$
This formulation regularizes the denoised latent rather than intermediate noisy states, incorporating physiological structure without waveform decoding to compute the energy term.

\vspace{-.3cm}
\paragraph{Energy-Regularized Training Objective}
Instead of optimizing a fully normalized energy-based model, we regularize the standard diffusion objective with our latent EPK energy:
$
\mathcal{L}
=
\mathbb{E}_{z_0,\epsilon,\tau}
\left[
\|\epsilon_\theta(z_\tau,\tau)-\epsilon\|_2^2
+
\lambda_{\text{EPK}}E_\phi(\hat{z}_0)
\right].
$
Conditioning arguments are omitted for brevity. Here, $\lambda_{\text{EPK}}$ is a fixed coefficient, with no additional timestep-dependent weighting. The main pharmacological-mapping experiments use $\lambda_{\text{EPK}}=0.1$. Default inference retains EPK conditioning and uses standard DDPM sampling without explicit energy-gradient correction.

\subsection{Uncertainty-Aware Risk Evaluation and Treatment Assessment}
\label{sec:risk}

Given the current state $z_k$, we evaluate a finite set of
candidate interventions $\mathcal{A}=\{a_1,\dots,a_M\}$,
where each action represents a discrete drug or regimen.
The framework supports VLM-based candidate proposal from
a specified intervention vocabulary.
In the reported experiments, we directly use fixed candidate
sets: eight predefined regimens for the interval and EPK
experiments, and the five most frequent medications in the
MIMIC-IV test split for the risk experiments.
The candidate sets and selection procedure are detailed
in Appendix~\ref{app:action_sets}.

To characterize variability across simulated outcomes,
we introduce an \textit{Uncertainty-Aware Evaluation Module}
that uses the stochastic generative nature of our model.
For each candidate intervention $a\in\mathcal{A}$, conditioned
on the current state $z_k$, we draw $S$ independent, fully
denoised next-state samples:
\begin{equation}
\hat{z}_{k+1}^{(s)}
\sim p_{\mathrm{WM}}(\cdot\mid z_k,a),
\qquad s=1,\dots,S,
\end{equation}
where $p_{\mathrm{WM}}$ denotes the trained conditional
world model sampled using standard DDPM with EPK conditioning.
A frozen ECG-FM diagnostic model $f_{\mathrm{risk}}$ evaluates
the decoded waveforms. Let $R(x)$ denote the scalar diagnostic
risk proxy obtained by aggregating its output probabilities.
Each generated sample is scored as
$
R^{(s)}(a)
=
R\!\left(f_{\mathrm{dec}}(\hat{z}_{k+1}^{(s)})\right).
$
The diagnostic labels and aggregation procedure are detailed
in Appendix~\ref{scr}.
We estimate the intervention-specific mean score and variance
across generated samples as
\begin{equation}
\mu_R(a)
=
\frac{1}{S}\sum_{s=1}^{S}R^{(s)}(a),
\qquad
\sigma_R^2(a)
=
\frac{1}{S-1}\sum_{s=1}^{S}
\left(R^{(s)}(a)-\mu_R(a)\right)^2.
\end{equation}
These quantities summarize the expected diagnostic-proxy
score under the model and its variability across stochastic
simulations.
We apply the risk-aware scoring function
$S(a)=\mu_R(a)+\lambda\sigma_R(a)$ introduced in
Equation~\ref{eq:final_decision_prelim}.
The coefficient $\lambda$ affects candidate ranking but
does not enter the diagnostic-change calculation for a
fixed recorded medication.
The resulting scores support comparative assessment within
the specified candidate set, rather than calibrated estimates
of clinical risk or treatment efficacy.

\subsection{Theoretical Foundations and Mechanistic Interpretation}
\label{sec:theory}

We interpret physiological regularization through an idealized energy-tilted distribution balancing a learned reference distribution with an ODE-derived anchor. This perspective motivates the practical training objective while distinguishing theoretical properties from empirical behavior. Detailed derivations are provided in Appendix~\ref{app:theory_proofs}.

\vspace{-.3cm}
\paragraph{Principled Optimality}
For a fixed reference density $p_\theta(z\mid a)$ and energy $E_\phi(z)$, the distribution $\tilde{p}(z\mid a)\propto p_\theta(z\mid a)\exp(-\gamma E_\phi(z))$ minimizes
\begin{equation}
\tilde{p}(\cdot\mid a)
=
\arg\min_q
\left\{
\mathrm{KL}\!\left(q \,\|\, p_\theta(\cdot\mid a)\right)
+
\gamma \mathbb{E}_q[E_\phi(z)]
\right\},
\end{equation}
assuming $\gamma>0$ and a finite, nonzero normalizing constant. The KL term favors agreement with the reference distribution, while the energy term penalizes deviations from the physiological anchor. This establishes a distribution-level trade-off between data fidelity and physiological preference, providing a rationale for regularization rather than an exact equivalence to the implemented training loss.

\vspace{-.3cm}
\paragraph{Score-Based Anchoring and Amortized Guidance}
At the clean-latent level, the idealized score admits an exact decomposition:
\begin{equation}
\nabla_z \log \tilde{p}(z\mid a)
=
\underbrace{\nabla_z \log p_\theta(z\mid a)}_{\text{Data-Driven Pull}}
-
\gamma
\underbrace{\nabla_z E_\phi(z)}_{\text{Prior-Driven Pull}}.
\end{equation}
This identity explains how physiological energy modifies the reference score, but does not directly specify intermediate noisy-state scores.
Our implementation incorporates physiological information through EPK conditioning and the training penalty $\lambda_{\text{EPK}}E_\phi(\hat{z}_0)$, which encourages predicted clean latents to remain near the projected ODE anchor. The energy term is evaluated directly in latent space without waveform decoding. Default inference uses standard DDPM sampling without explicit energy-gradient correction, with physiological guidance supplied through conditioning and the regularized model parameters.

Under suitable regularity and boundary conditions, the Fokker--Planck analysis establishes $\tilde{p}$ as stationary for the specified continuous-time Langevin dynamics. This result does not establish exact sampling by the deployed DDPM or automatic fallback under prior mismatch. Robustness is assessed empirically through generated-sample dispersion under prior perturbations and short-chain terminal errors (Appendices~\ref{b9} and \ref{e5}), rather than inferred from a theoretical safety guarantee.

\vspace{-.3cm}
\section{Experiments}
\vspace{-.2cm}
\label{sec:experiments}

\subsection{Experiment Setup}

\paragraph{Datasets}

We evaluate our framework using a combination of real-world clinical electronic health records (EHRs) and controlled ECG datasets. To strictly prevent data leakage, each dataset is partitioned at the subject level based on patient IDs (about 8:2 train-test split) (more details in Appendix \ref{dataDescription}):  (1) \textbf{MIMIC-IV-ECG \& Clinical} -- By integrating MIMIC-IV-ECG \citep{gow2023mimic} with MIMIC-IV-Clinical \citep{johnson2023mimic}, we align 12-lead ECG recordings with longitudinal patient EHRs. This enables multi-modal conditioning on demographics, lab results, and existing comorbidities. The dataset comprises 17,493 unique patients and 108,980 patient–drug episodes, with 3,090 distinct medications;  (2) \textbf{Drug-Response ECG Datasets} -- To evaluate drug-induced cardiac morphological effects, we utilize the ECGRDVQ \citep{johannesen2014differentiating} and ECGDMMLD \citep{johannesen2016late} databases. These provide 7,763 valid high-fidelity ECGs from 43 subjects across eight distinct drug regimens; and %Following quality control and VAE latent extraction, 7,763 valid samples from 43 subjects are retained. Specifically, 34 subjects comprising 6,143 samples are assigned to the training set, and the remaining 9 subjects with 1,620 samples to the test set. This rigorous subject-disjoint division serves as the standard for assessing the physiological fidelity. 
(3) \textbf{In-house OOD Dataset} -- To assess cross-dataset generalization, we test our model, trained primarily on the controlled drug-response cohorts, against an in-house dataset of MIMIC-IV ICU patients administered Verapamil. %This OOD setting evaluates the model's robustness to real-world clinical noise and diverse patient populations.

%(1) \textbf{MIMIC-IV-ECG} Integrating MIMIC-IV-ECG \citep{gow2023mimic} and MIMIC-IV-Clinical \citep{johnson2023mimic} aligns 12-lead ECGs with patient EHRs; and  (2) \textbf{Drug-Response ECG Datasets} To evaluate drug-induced cardiac morphological effects, we use the ECGRDVQ \citep{johannesen2014differentiating} and ECGDMMLD \citep{johannesen2016late} databases. %\textbf{Drug-Response ECG Datasets.} To evaluate drug-induced cardiac morphological effects, we use the ECGRDVQ \citep{johannesen2014differentiating} and ECGDMMLD \citep{johannesen2016late} databases. 
%$These provide an initial 9,443 ECGs from 44 subjects across eight drug regimens. After quality control, successful VAE latent extraction, and alignment with labels, 7,763 valid samples from 43 subjects are utilized. 
%Furthermore, \textbf{In-house OOD Evaluation Dataset}  assesses cross-dataset generalization, we test our model, trained on the healthy drug-response ECGs, against an in-house dataset of MIMIC-IV ICU patients administered Verapamil. 

% \textbf{Simulation-Driven Clinical Evaluation.} We leverage our world model strictly for inference-time predictive simulation. Given a latent state and candidate treatments, it simulates plausible post-treatment ECG trajectories. To prevent data leakage, these simulations are evaluated by a frozen ECG foundation model \citep{mckeen2025ecg}. Acting as an unbiased, zero-shot risk estimator, it ensures our uncertainty-aware framework accurately plans safe interventions without overfitting.

\vspace{-.3cm}
\paragraph{Training Details}
Our world model is built upon a Denoising Diffusion Probabilistic Model (DDPM) backbone \citep{ho2020denoising}, augmented with an action-conditioned module to model intervention effects. During training, we incorporate an ECG-specific ODE prior as a physiological constraint to guide the diffusion process and capture the underlying cardiac dynamics. We train the model using AdamW with a learning rate of $5\times10^{-4}$ and a batch size of 64 for 200 epochs. A cosine annealing scheduler is applied to adjust the learning rate during training. The main pharmacological-mapping experiments use
an energy coefficient $\lambda_{\text{EPK}}=0.1$.
The energy-weight sweep in Appendix~\ref{app:energy_weight_ablation} retains
its separately reported coefficients, including 0.25.
Diagnostic-score evaluation uses $S=5$ stochastic
samples per candidate intervention.
Default inference uses standard DDPM sampling,
denoted by $K_{\mathrm{res}}=1$.
The $K_{\mathrm{res}}=8$ configuration is evaluated
 in the separate 48-pair pilot in Appendix~\ref{e6}.
The action-selection coefficient $\lambda$ is
reported for each selection evaluation.

%%%%% 1. 药量的问题，目前的数据集中没有关于药量的指标，因此这种可以结合不同drug的医疗说明来将这个打入进去
%%%%% 2. risk prediction model的问题，目前关于风险评估的模型没有一个很好的，能够彻底进行训练的模型可以进行
%%%%% 3. 可以借助于openclaw来进行数据的生成，从而来评估world model带来知识的好坏

\vspace{-.3cm}
\paragraph{Architecture Implementation}

We adopt a modular design to implement the proposed framework, where each component is instantiated using either pre-trained foundation models or task-specific adaptations: (1) Treatment Action Interface -- The framework supports
VLM-based candidate proposal (GPT-4o \citep{hurst2024gpt}, DeepSeek-VL \citep{lu2024deepseekvl}, BLIP-FlanT5 \citep{li2023blip}), while the reported experiments
directly evaluate the fixed intervention sets specified in
Appendix~\ref{app:action_sets}; (2) Physiology-Informed World Model --
We model cardiac dynamics using our proposed latent diffusion world model. To ensure physiological consistency and mitigate hallucinations, External Physiological Knowledge (EPK) from cardiac ODEs is injected as both conditioning signals and an energy-based regularization term during the denoising process; (3) Variational Autoencoder (VAE) --
To obtain compact representations of the 12-lead ECGs, we use a VAE pre-trained  \citep{lai2025diffusets} on MIMIC-IV-ECG. We keep this module completely frozen to maintain a consistent latent space across all experiments; and (4) Downstream Risk Prediction Model --
For risk evaluation, we use a state-of-the-art ECG foundation model \citep{mckeen2025ecg}, and this model is kept strictly frozen to assess simulated post-dose ECGs in a purely zero-shot manner.

\vspace{-.3cm}
\paragraph{Baselines}
%\vspace{-.2cm}

To rigorously evaluate our framework, we conduct comprehensive comparisons across two major dimensions: physiological simulation fidelity and downstream clinical evaluation.

\textbf{(1) ECG Generative and Dynamics Models}
To evaluate \textit{waveform quality and physiological consistency} during simulation, we include a wide range of open-loop generative models for ECG signals, covering both GAN-based and diffusion-based approaches. Specifically, we consider classical GAN variants (\textit{WGAN} \citep{arjovsky2017wasserstein}, \textit{StyleGAN} \citep{karras2019style}, \textit{ECG ODE-GAN} \citep{golany2021ecg}, \textit{TTS-CGAN} \citep{li2022tts}, and \textit{CECG-GAN} \citep{yang2024data}), as well as recent  diffusion-based models (\textit{Diffusets} \citep{lai2025diffusets}, \textit{SE-Diff} \citep{wang2026se} and \textit{DADM} \citep{shao2025generation}).  We utilize these approaches as purely data-driven baselines to highlight the effectiveness of our proposed method in mitigating generative inconsistencies.

\textbf{(2) Intervention-Aware Simulation and Comparative Evaluation}iv
%Moving beyond unconditional generation, we evaluate the model's ability to simulate ECG dynamics under structured pharmacological interventions, including \textit{single-step prediction}, \textit{multi-step rollouts}, and \textit{out-of-distribution (OOD) interventional scenarios}. For the downstream \textit{Intervention Ranking Consistency} task, we compare our simulation-driven approach against representative large-scale clinical foundation models and multi-modal LLMs, including \textit{Qwen2.5-VL-7B} \citep{bai2025qwen2}, \textit{GPT-series} \citep{achiam2023gpt}, and \textit{MedGemma} \citep{sellergren2025medgemma}. These models serve as strong zero-shot baselines that attempt to infer intervention preferences directly from the context. 
Moving beyond unconditional generation, we evaluate action-conditioned dynamics across \textit{single-step prediction}, \textit{multi-step rollouts}, and \textit{out-of-distribution (OOD) scenarios}. For the downstream \textit{Intervention Ranking Consistency} task, we compare our simulation-driven approach against representative large-scale clinical foundation models and multi-modal LLMs, including \textit{Qwen2.5-VL-7B} \citep{bai2025qwen2}, \textit{GPT-series} \citep{achiam2023gpt}, \textit{GLM-series} \citep{glm2024chatglm}, and \textit{MedGemma} \citep{sellergren2025medgemma}. These models serve as strong zero-shot baselines that attempt to infer intervention preferences directly from clinical context, rather than through explicit physiological simulation.

% \vspace{-.8cm}
\subsection{Experimental Results}
\label{sec:res_mimic_emergence}

\begin{wraptable}{l}{0.5\textwidth}
\vspace{-.8cm} 

% 标题部分保持你现在的写法不变
\caption{\parbox[t]{0.8\linewidth}{\raggedright Comparison of methods on latent and signal Post-dose ECG prediction.}}
\label{tab:comparison1}

% --- 核心修改：用 resizebox 把表格强制缩放到包裹盒子的宽度 ---
\resizebox{\linewidth}{!}{
\begin{tabular}{l|cc|cc}
\hline
\multirow{2}{*}{Methods} 
& \multicolumn{2}{c|}{Latent ECG} 
& \multicolumn{2}{c}{Signal ECG}  \\
& MSE $\downarrow$ & MAE $\downarrow$ 
& MSE $\downarrow$ & MAE $\downarrow$ \\
\hline
Qwen2.5-VL-7B & 2.148 & 0.368 
& 0.195 & 0.201  \\
GPT-5 mini & 0.719 & 0.166
& 0.066 & 0.128 \\
GPT-4o & 0.550 & 0.414
& 0.578 &  0.386\\
GPT-3.5 & 0.424 & 0.285
& 0.265 & 0.232  \\
GLM-4.5& 0.095 & 0.164
& 0.062 & 0.125  \\
MedGemma & 5.655	& 0.899 
& 0.561 & 0.349  \\
\rowcolor{blue!8}
Ours & \textbf{0.045} & \textbf{0.159} 
& \textbf{0.053} & \textbf{0.110}  \\
\hline
\end{tabular}
} % <--- 注意这里有个大括号的闭合，是 resizebox 的结束
\vspace{-.3cm} 
\end{wraptable}

\paragraph{Evaluation of Emergent Capabilities on Clinical Dynamics} 
We investigate whether the emergent capabilities of general-purpose foundation models extend to continuous cardiac dynamics. %As shown in Table~\ref{tab:comparison1}, despite their strong performance in language and vision tasks, these models exhibit notable limitations in modeling ECG dynamics, resulting in higher errors in both latent physiological space and signal reconstruction on the MIMIC-IV-ECG benchmark. 
Since this cohort primarily consists of abnormal cases, the ECG signals often lack the regular patterns observed in normal populations. Under this setting, we evaluate model performance using mean squared error (MSE) and mean absolute error (MAE). 
As shown in Table~\ref{tab:comparison1}, despite their strong reasoning performance in language and vision tasks, these zero-shot models exhibit notable limitations in modeling fine-grained ECG dynamics. This results in substantially higher prediction errors in both the latent physiological space and the decoded signal reconstruction on the MIMIC-IV-ECG benchmark.

\begin{table*}[h!]
\vspace{-.5cm}
\centering
\caption{Performance comparison of different methods on three ECG intervals.}
\resizebox{0.98\linewidth}{!}{
\begin{tabular}{l|cc|cc|cc}
\hline
\multirow{2}{*}{Methods} 
& \multicolumn{2}{c|}{$QT_c$ Interval} 
& \multicolumn{2}{c|}{$PR$ Interval} 
& \multicolumn{2}{c}{$T_\text{peak}-T_\text{end}$ Interval} \\
& Acc (\%) $\uparrow$ & Rec (\%) $\uparrow$ 
& Acc (\%) $\uparrow$ & Rec (\%) $\uparrow$
& Acc (\%) $\uparrow$ & Rec (\%) $\uparrow$ \\
\hline

WGAN & $75.23_{\pm 0.69}$ & $77.23_{\pm 0.71}$  & $57.67_{\pm 0.64}$ & $58.04_{\pm 0.54}$  & $85.49_{\pm 0.12}$ & $87.35_{\pm 0.16}$ \\
StyleGAN & $72.71_{\pm 1.06}$ & $74.29_{\pm 1.49}$ & $52.15_{\pm 3.54}$ & $55.35_{\pm 1.20}$ & $88.71_{\pm 1.96}$ & $90.35_{\pm 1.87}$ \\
ECG ODE-GAN &  $72.81_{\pm 3.24}$ & $73.95_{\pm 2.25}$ & $54.63_{\pm 1.49}$ & $56.16_{\pm 1.39}$ & $87.03_{\pm 0.52}$ & $88.19_{\pm 0.79}$ \\
TTS-CGAN & $86.43_{\pm 1.65}$ & $87.21_{\pm 0.64}$ & $70.09_{\pm 2.08}$ & $72.82_{\pm 2.34}$ & $83.43_{\pm 1.85}$ & $85.48_{\pm 1.57}$ \\
CECG-GAN &$79.92_{\pm 1.27}$ & $82.75_{\pm 1.31}$ & $61.08_{\pm 1.43}$ & $60.62_{\pm 0.71}$ & $89.29_{\pm 0.13}$ & $91.26_{\pm 0.05}$ \\

% DiffECG & 83.48 & 77.50 & 83.65 & 82.35 & 85.86 & 85.45 \\
% BioDiffusion & 80.41 & 72.50 & 85.01 & 79.41 & 83.82 & 83.64 \\

Diffusets & $86.81_{\pm 0.98}$ & $90.07_{\pm 1.24}$ &$72.62_{\pm 1.34}$ & $74.20_{\pm 1.54}$ & $89.84_{\pm 0.98}$ & $92.10_{\pm 0.96}$ \\
SE-Diff & $89.19_{\pm 0.58}$ & $89.19_{\pm 0.58}$ & $70.83_{\pm 0.42}$ & $73.58_{\pm 0.66}$ & $92.35_{\pm 0.78}$ & $94.22_{\pm 0.81}$ \\
DADM & $89.46_{\pm 1.36}$ & $93.69_{\pm 1.49}$ & $68.02_{\pm 1.14}$ & $72.26_{\pm 1.72}$ &$91.63_{\pm 0.76}$ & $93.04_{\pm 0.58}$\\
\hline
\rowcolor{blue!8}
Ours & $\textbf{90.55}_{\pm 0.76}$ & $\textbf{96.26}_{\pm 0.82}$  
& $\textbf{76.04}_{\pm 0.97}$ & $\textbf{80.84}_{\pm 1.24}$ 
& $\textbf{94.70}_{\pm 1.13}$ & $\textbf{95.06}_{\pm 0.49}$ \\
\hline

\end{tabular}
}
\label{tab:comparison}
\vspace{-.3cm}
\end{table*}

\paragraph{Physiological Simulation Fidelity} 

For normal cohorts, we adopt three evaluation metrics to provide a more comprehensive assessment.
As shown in Table~\ref{tab:comparison}, conventional metrics such as MSE are insufficient to capture clinically relevant characteristics of physiological signals. Instead, we evaluate models based on their ability to preserve key cardiac biomarkers, including $QT_c$, $PR$, and $T_{\text{peak}}-T_{\text{end}}$ intervals. A prediction is considered correct only if all indicator labels match the ground truth. Following \citep{shao2025generation}, $QT_c$ is computed using the Bazett's formula $QT_c = QT / \sqrt{RR}$, with normal ranges defined as $\leq 450$ ms for men and $\leq 470$ ms for women. The $PR$ interval is considered normal within $120$–$200$ ms, while $T_{\text{peak}}-T_{\text{end}}$ is normal within $80$–$113$ ms. Under this clinically motivated criterion, our method more accurately preserves physiologically consistent signal characteristics compared to all baselines, particularly under drug-induced variations.

\begin{wraptable}{l}{0.45\textwidth}
\vspace{-0.9cm}

\centering % 让表格内容在 wraptable 内居中

\caption{\small Out-of-distribution (OOD) comparison across different methods.}
\label{tab:model_comparisono}

{\small % 控制整体字体大小（关键！！）
\begin{tabular}{lcc}
\toprule
\textbf{Models} & \textbf{Latent} & \textbf{Signal}\\
\midrule
Diffusets & 0.096 & 0.150\\
DADM & 0.080 & 0.119\\
\rowcolor{blue!8}
Ours & \textbf{0.068} & \textbf{0.084} \\
\bottomrule
\end{tabular}
}

\vspace{-0.3cm}
\end{wraptable}

\paragraph{OOD Generalization on Unseen Diseases} 
%As shown in Figure~\ref{fig:ood}, we evaluate out-of-distribution (OOD) generalization against strong diffusion-based baselines (DiffuSETS and DADM) by training on the \textit{Drug-Response ECG dataset} and testing on an \textit{in-house evaluation dataset} containing unseen comorbidities and disease conditions. This setting reflects real-world deployment, where models must generalize to previously unseen pathological patterns.

As shown in Table~\ref{tab:model_comparisono}, we evaluate out-of-distribution (OOD) generalization against strong diffusion-based baselines (DiffuSETS and DADM). To rigorously test this, all models are trained exclusively on the controlled \textit{Drug-Response ECG Datasets} and evaluated on the \textit{In-house OOD Dataset}, which contains patients with unseen comorbidities and complex underlying pathologies. This setting reflects real-world clinical deployment, where generative models must safely generalize to previously unseen pathological patterns without hallucinating physically impossible cardiac dynamics.

% As shown in Figure~\ref{fig:ood}, we compare our method with strong diffusion-based baselines (DiffuSETS and DADM) and include ablations of our physiological components. Data-driven baselines exhibit significant performance degradation under distribution shift, while our model remains robust. We observe consistent improvements as physiological constraints are progressively introduced. Using EPK as input conditioning already improves performance, and further incorporating \textit{EPK-guided Energy Regularization} yields the best results across both latent and signal spaces. These results suggest that embedding physiological priors provides a robust inductive bias, enabling stable generalization to unseen clinical conditions.

% \begin{table*}[h!]
% \centering
% \caption{Performance comparison of unseen disease.}
% % \resizebox{\linewidth}{!}{
% \begin{tabular}{l|c|c}
% \hline
% Methods
% & Latent ECG
% & Signal ECG  \\
% \hline
% Diffusets &  0.096
% & 0.150    \\
% DADM & 0.080 
% & 0.119  \\
% Ours & 0.068 
% & 0.084   \\
% \hline
% \end{tabular}
% \label{tab:comparison}
% \end{table*}

\paragraph{Drug-Induced Risk Modeling and Directional Consistency} 
%To evaluate the effectiveness of our predictive simulation framework, we focus on modeling the \textit{relative risk change} ($\Delta$Risk) between pre-dose and post-dose states, rather than static risk prediction. 
\vspace{-.3cm} 
\begin{wraptable}{l}{0.5\textwidth}
\vspace{-.4cm} 

\caption{\parbox[t]{0.82\linewidth}{\raggedright Comparison of Core Evaluation Metrics between Models}}
\label{tab:model_comparison}

\resizebox{\linewidth}{!}{
\begin{tabular}{l>{\columncolor{blue!8}}cc}
\toprule
\textbf{Core Evaluation Metric} & \textbf{Ours} & \textbf{DADM}\\
\midrule
$\Delta$Risk Pearson Correlation & \textbf{0.620} & 0.266\\
$\Delta$Risk Spearman Correlation & \textbf{0.598} & 0.255\\
$\Delta$Risk Sign Agreement & \textbf{76\%} & 58\% \\
MAE & \textbf{0.0973} & 0.1291\\
RMSE & \textbf{0.1253} & 0.1625 \\
\bottomrule
\end{tabular}
}

\vspace{-0.3cm}
\end{wraptable}

We evaluate diagnostic-change fidelity on 200 subject-disjoint
MIMIC-IV test cases covering Propofol, Regular Insulin,
Prophylactic Heparin Sodium, Furosemide, and Norepinephrine.
Diagnostic scores average 17 label probabilities from a frozen
ECG-FM. For each recorded medication, predicted post-dose scores
are averaged over $S=5$ standard-DDPM samples; predicted and
observed changes use the same pre-dose reference.
Ours improves agreement and reduces errors over DADM
(Table~\ref{tab:model_comparison}), achieving Pearson $r=0.620$,
Spearman $\rho=0.598$, and 76\% sign agreement (rounded)
in the reported run.
Score definitions and all six sampling runs are provided in
Appendix~\ref{scr} and Table~\ref{tab:diagnostic_change_runs},
respectively.

\textbf{Additional Analyses in Appendix}
We provide additional analyses of
(i) EPK-module ablation and short-chain rollout--oracle
comparisons (Figures~\ref{fig:ab1} and \ref{fig:ab2},
and Table~\ref{tab:rollout_comparison});
(ii) diagnostic-change agreement across repeated stochastic
sampling runs (Figure~\ref{fig:ab3}) and the number of
stochastic samples $K$ (Appendix~\ref{sec:res_ablationk});
(iii) the energy coefficient $\lambda_{\text{EPK}}$
with EPK conditioning retained
(Table~\ref{tab:energy_weight_ablation});
(iv) reconstruction under missing-lead settings in normal
(Table~\ref{tab:missing_leads_old_new} and
Figure~\ref{fig:v12}) and abnormal conditions
(Figure~\ref{fig:v22});
and (v) generated-sample variability and latent dispersion
under prior perturbations (Figure~\ref{fig:v32}),
where EPK reduces sample dispersion across the evaluated
perturbation regimes.

\vspace{-.4cm}
\section{Conclusion}
\vspace{-.3cm}
%%前60%写conclusion，后40%写limitation（最少两个点）
%%% 我们从一个新的视角探索了ERM方法以及IRM-based方法失效的内生性原因，是因为语义特征在跨环境时，无法进行对齐，从而导致了泛化性能的降低。我们使用神经坍塌现象来指导跨环境的条件下，语义特征的对齐过程，显著提升了OOD性能。同时，考虑到环境标签未知的情况，我们可以自己进行环境的划分，这大大增加了该方法的适用范围。我们的局限性可以在后续的工作中进行更加effective and efficient的划分环境的操作 
In this work, we propose an ECG World Model for intervention-aware cardiac analysis, enabling simulation-driven modeling of drug-induced dynamics. By integrating intervention generation, physiology-informed latent dynamics, and risk evaluation, our framework extends ECG analysis from static prediction toward intervention-conditioned simulation. Experiments demonstrate strong performance in modeling drug responses and generalization across multiple clinical datasets. Several limitations remain. First, although the framework supports continuous dosage representations, current public datasets lack sufficiently dense pharmacokinetic measurements, so we use discrete intervention tokens. Second, the physiology-guided energy regularization relies on classical ODE priors, which may be less suitable for highly irregular rhythms such as severe ventricular fibrillation. Third, the constrained action space provides only an approximation of complex ICU pharmacology and does not fully capture polypharmacy interactions. Finally, the current results are proof-of-concept and require substantially broader clinical validation before real-world use. 

% \section*{Acknowledgement}
% This work was supported in part by the National Natural Science Foundation of China (No. 623B2002) and the National Natural Science Foundation of China (No. 62176132). MG was supported by ARC DE210101624, ARC DP240102088, and WIS-MBZUAI 142571. Sen Cui would like to acknowledge the financial support received from Shuimu Tsinghua scholar program.

\bibliography{iclr2027_conference}
\bibliographystyle{iclr2027_conference}

\newpage
% \section{Implementation Details}
% \subsection{Datasets}
% \label{ap:datasets}

% \section{Complete Mathematical Derivation of the ECG World Model}
% \label{app:theory_complete}

\appendix

\section{Datasets}\label{dataDescription}

We utilize a combination of real-world clinical electronic health records and controlled ECG datasets to comprehensively evaluate our framework.

\paragraph{MIMIC-IV-ECG \& Clinical Cohort}
Our primary dataset is constructed by integrating the MIMIC-IV-ECG \citep{gow2023mimic} and MIMIC-IV-Clinical \citep{johnson2023mimic} databases. MIMIC-IV-ECG contains a large collection of 12-lead electrocardiogram (ECG) waveform recordings, while MIMIC-IV-Clinical provides comprehensive electronic health record (EHR) data, including patient demographics, diagnoses, laboratory measurements, procedures, and clinical outcomes. We link the two datasets through shared patient identifiers, enabling the alignment of ECG signals with corresponding clinical context. In particular, medication records are aligned with each patient and their corresponding pre-dose ECG observations, forming patient–drug pairs that reflect real-world treatment trajectories. The ECG data capture fine-grained cardiac dynamics, while explicit exposure to drugs allows the world model to learn patient-specific state transitions. This multimodal dataset serves as our primary testbed for evaluating \textit{treatment consistency}, allowing us to compare model-inferred intervention preferences with clinician-prescribed treatments. To prevent data leakage, we perform dataset splitting at the patient level based on unique subject identifiers, ensuring that no patient appears in both training and test sets. The cohort includes 17,493 unique patients, of which 13,994 are assigned to the training set and 3,499 to the test set. At the episode level, the dataset comprises 108,980 samples in total, with 87,652 training episodes and 21,328 test episodes, consistent with the total number of processed records. In addition, the dataset includes 3,090 distinct medications, providing a diverse set of intervention signals for modeling drug-induced physiological responses.

\paragraph{Drug-Response ECG Datasets.}
To evaluate drug-induced ECG changes, we use the ECGRDVQ
database
and the ECGDMMLD database.
The experimental cohort used in this study comprises 7,763
12-lead ECG recordings from 43 subjects across eight drug regimens:
Dofetilide, Lidocaine+Dofetilide, Mexiletine+Dofetilide,
Moxifloxacin+Diltiazem, Placebo, Quinidine, Ranolazine,
and Verapamil.
We partition this cohort at the subject level into 6,143 recordings
from 34 subjects for training and 1,620 recordings from 9 subjects
for testing.
No subject appears in both splits.
These datasets are used to evaluate the fidelity of
action-conditioned ECG generation.

\paragraph{In-house OOD Evaluation Dataset}
To further evaluate the effectiveness of our approach in realistic clinical settings, we adopt a cross-dataset generalization strategy. Specifically, our model is trained on Drug-response ECG datasets, which consist of ECG recordings from healthy individuals following drug administration. To assess the model’s ability to generalize to more clinically relevant and challenging out-of-distribution (OOD) scenarios, we construct an in-house evaluation dataset based on MIMIC-IV-ECG. This dataset contains ECG recordings from ICU patients after the administration of Verapamil, reflecting real-world clinical conditions with underlying diseases and complex physiological variability. Therefore, this dataset provides a rigorous benchmark for evaluating the robustness of our physiology-guided energy regularization against severe distribution shifts.

%\textbf{Simulation-Driven Clinical Evaluation}
%Instead of using generated data for training augmentation or fine-tuning, we leverage the proposed world model strictly for \textit{inference-time predictive simulation}. Given a latent cardiac state and candidate interventions, the model simulates multiple plausible post-intervention ECG trajectories. To avoid circular validation or data leakage, these simulated trajectories are evaluated using a frozen ECG foundation model \citep{mckeen2025ecg}, which serves as an objective zero-shot risk estimator. This evaluation protocol ensures that our uncertainty-aware assessment reflects the model’s ability to capture intervention-dependent physiological dynamics, rather than overfitting to synthetic distributions.

% \section{Robust ECG Reconstruction under Missing-Lead Settings}

% Figures~\ref{fig:vv1},~\ref{fig:vv2},~\ref{fig:vv3} and~\ref{fig:vv4}  present qualitative results under various missing-lead patterns. Despite incomplete inputs, our model reconstructs coherent ECG signals and preserves key waveform structures across the leads. Compared to typical data-driven approaches, the reconstructed signals remain physiologically consistent, without introducing noticeable artifacts or distortions. This shows that incorporating physiological priors enables robust recovery of cardiac dynamics even under severe missing conditions.

\section{Experiments}

\subsection{Ablation Studies: The Role of Energy Regularization}
\label{sec:res_ablation}

To rigorously isolate the contribution of our proposed physiological constraints, we conduct an ablation study comparing the full ECG World Model against an unconstrained diffusion variant (i.e., removing the EPK-informed Energy Regularization).

\begin{figure*}[h!]
    % \vspace{-.1cm}
    % \setlength{\abovecaptionskip}{-10cm}
    % \setlength{\belowcaptionskip}{-10cm}
    \centering
    \includegraphics[width=1\columnwidth]{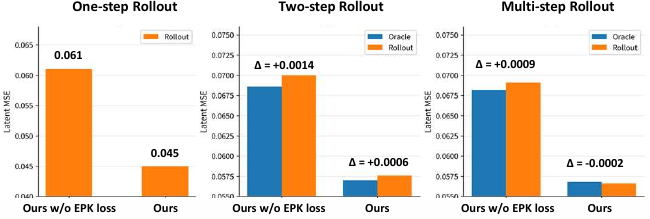}
    \vspace{-0.6cm}
    \caption{The overview of rollout settings used in this study. One-step rollout evaluates single-step prediction on the Drug-response ECG dataset, while two-step and multi-step rollouts assess short- and long-horizon predictions on the MIMIC-IV-ECG dataset.}
    \label{fig:ab1}
    % \vspace{-.6cm}
\end{figure*}

\paragraph{Mitigating Error Accumulation in Multi-Step Rollouts}
As illustrated in Figure~\ref{fig:ab1}, our model consistently outperforms the unconstrained variant across all rollout horizons. In the one-step setting, energy regularization reduces the latent MSE from 0.061 to 0.045. As the rollout horizon increases, the unconstrained model exhibits significant error accumulation and deviates from physiologically plausible cardiac trajectories, while our method maintains stable predictions. 
This effect is further confirmed in Table~\ref{tab:rollout_comparison}, where we compare autoregressive rollout performance with oracle predictions. The gap ($\Delta$) between rollout and oracle errors remains minimal for our model (+0.0008), whereas general-purpose models exhibit substantially larger deviations. These results suggest that our approach effectively mitigates temporal error accumulation and improves long-horizon consistency.

\begin{table}[h!]
\centering
% \caption{Reconstruction performance under missing-lead settings. Comparison between Ours and Ours wo EPK loss test sets using 50 samples (lower is better).}
% \label{tab:missing_leads_old_new}
% \resizebox{\linewidth}{!}{

% 标题部分：使用 parbox 限制宽度并靠左对齐
\caption{ Comparison of autoregressive multi-step rollout and oracle predictions across models using MIMIC-IV-ECG dataset. $\Delta$ denotes the difference between rollout and oracle latent signal mse.}
% 建议加个 label 方便正文引用
\label{tab:rollout_comparison} 

% 核心修改：用 resizebox 把表格强制缩放到包裹盒子的宽度，防止越界重叠
% \resizebox{\linewidth}{!}{
\begin{tabular}{lccc}
\hline
Model & Oracle $\downarrow$ & Rollout $\downarrow$ & $\Delta$ \\
\hline
GPT-4o        & 0.1124  & 0.1245  & +0.0121    \\
GPT-5-mini    & 0.1025 & 0.1172 & +0.0147  \\
GPT-3.5       & 0.0942 & 0.1036 & +0.0094  \\
% Ours wo EPK loss    & 0.0734 & 0.0734 & \textbf{0.0000}   \\
\rowcolor{blue!8}
Ours  & \textbf{0.0572} & \textbf{0.0580} & \textbf{+0.0008}  \\
\hline
\end{tabular} % <--- resizebox 的结束括号

\vspace{-.3cm}
\end{table}

\begin{figure*}[h!]
    % \vspace{-.1cm}
    % \setlength{\abovecaptionskip}{-10cm}
    % \setlength{\belowcaptionskip}{-10cm}
    \centering
    \includegraphics[width=1\columnwidth]{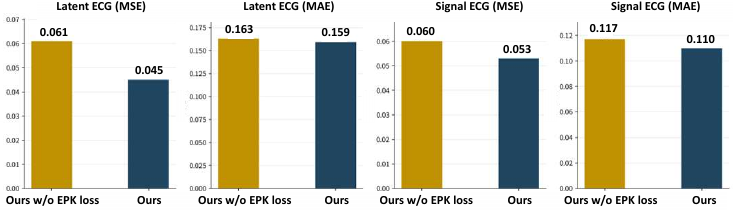}
    \vspace{-0.6cm}
    \caption{Ablation study on the Drug-response ECG dataset. We compare our full model with a variant without EPK loss across both latent and signal spaces using MSE and MAE metrics, demonstrating the effectiveness of the EPK loss in improving prediction quality.}
    \label{fig:ab2}
    \vspace{-.5cm}
\end{figure*}

\paragraph{Fidelity Across Latent and Signal Spaces} 
Figure~\ref{fig:ab2} further shows that removing the energy regularization mechanism degrades predictive fidelity in both the latent space and the high-dimensional signal space. Our fully regularized model achieves consistently lower MSE and MAE across both domains. 
These ablation results support our core premise that purely data-driven diffusion models are prone to temporal inconsistencies during multi-step intervention simulation. Incorporating physiological priors via energy regularization plays a key role in improving temporal consistency and predictive stability.

% \begin{table}[t!]
% \centering
% \caption{Reconstruction performance on latent and signal ECG (lower is better).}
% \label{tab:latent_signal_mse_mae}
% % \resizebox{\linewidth}{!}{
% \begin{tabular}{l|cc|cc}
% \hline
% \multirow{2}{*}{Methods} 
% & \multicolumn{2}{c|}{Latent ECG} 
% & \multicolumn{2}{c}{Signal ECG} \\
% & MSE $\downarrow$ & MAE $\downarrow$ 
% & MSE $\downarrow$ & MAE $\downarrow$ \\
% \hline
% missL0  & 0.0451 & 0.1650 & 0.0413 & 0.1013 \\
% missL1  & 0.0454 & 0.1650 & 0.0420 & 0.1017 \\
% missL2  & 0.0449 & 0.1644 & 0.0405 & 0.1010 \\
% missL3  & 0.0453 & 0.1641 & 0.0399 & 0.0999 \\
% missL4  & 0.0426 & 0.1602 & 0.0357 & 0.0946 \\
% missL5  & 0.0443 & 0.1632 & 0.0381 & 0.0984 \\
% missL6  & 0.0442 & 0.1625 & 0.0405 & 0.0999 \\
% missL7  & 0.0458 & 0.1653 & 0.0412 & 0.1016 \\
% missL8  & 0.0436 & 0.1611 & 0.0404 & 0.0991 \\
% missL9  & 0.0433 & 0.1607 & 0.0395 & 0.0972 \\
% missL10 & 0.0442 & 0.1622 & 0.0421 & 0.0995 \\
% missL11 & 0.0445 & 0.1629 & 0.0401 & 0.0979 \\
% \hline
% \end{tabular}
% \end{table}

\subsection{Repeated-Sampling Evaluation of Diagnostic-Change Fidelity}
\label{sec:res_ablation1}

We evaluate diagnostic-change fidelity across six stochastic
sampling runs on the same 200 test cases.
For each case, the predicted post-dose diagnostic score for
the recorded medication is averaged over $S=5$ standard-DDPM
samples. Predicted and observed diagnostic changes are
computed relative to the same pre-dose reference using the
frozen diagnostic backend.

The risk-aversion coefficient $\lambda$ is used only for
candidate selection through the score
$\mu_R(a)+\lambda\sigma_R(a)$.
It does not enter the diagnostic-change calculation for
the recorded medication. Although the six evaluation files
were associated with different selection coefficients,
each run regenerated its DDPM samples.
Differences in the reported correlations and sign agreement
therefore reflect stochastic sampling variation rather
than an isolated effect of $\lambda$.

\begin{figure*}[h!]
    \centering
    % Update the horizontal tick labels in abn.pdf to
    % Run 1, Run 2, ..., Run 6; retain all numerical values.
    \includegraphics[width=0.8\columnwidth]
    {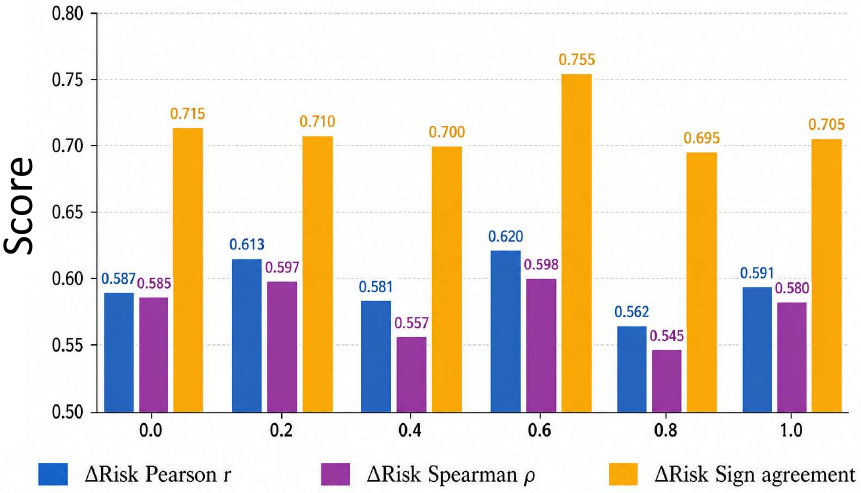}
    \caption{
    Diagnostic-change agreement across six stochastic
    sampling runs on the same 200 test cases.
    Each run regenerates standard-DDPM samples for the
    recorded medication. Bars report Pearson $r$,
    Spearman $\rho$, and sign agreement between predicted
    and observed diagnostic-score changes.
    The horizontal axis indexes sampling runs, not a
    risk-aversion ablation: $\lambda$ is used only for
    candidate selection and does not enter these
    diagnostic-change metrics.
    }
    \label{fig:ab3}
\end{figure*}

As shown in Figure~\ref{fig:ab3}, positive diagnostic-change
agreement is observed across all six runs.
Pearson correlation is $0.592\pm0.021$, Spearman correlation
is $0.577\pm0.022$, and sign agreement is $(71.3\pm2.2)\%$.
These values report the mean and sample standard deviation
across the six runs, quantifying evaluation-sampling
variability rather than training-seed variability.
The results support diagnostic-change fidelity across
repeated stochastic sampling under the frozen evaluation
backend. Complete run-level results and the evaluation
definitions are provided in Appendix~\ref{e4}.

\subsection{Architecture-Preserving Energy-Weight Ablation}
\label{app:energy_weight_ablation}

We vary the physiological energy coefficient
$\lambda_{\text{EPK}}$ while retaining the EPK architecture
and conditioning pathway.
In particular, $\lambda_{\text{EPK}}=0$ removes the energy
penalty but does not disable EPK conditioning.
This setting is therefore distinct from the
EPK-module removal in Figure~\ref{fig:ab2}.

\begin{table}[t]
\centering
\small
\caption{
Architecture-preserving ablation of the physiological
energy coefficient $\lambda_{\text{EPK}}$.
EPK conditioning remains enabled in every row;
only the energy-loss coefficient is varied.
The coefficients and numerical results are retained
from this weight-sweep protocol, including the
$\lambda_{\text{EPK}}=0.25$ setting.
}
\label{tab:energy_weight_ablation}
\begin{tabular}{c|cc|cc}
\toprule
& \multicolumn{2}{c|}{Latent Space}
& \multicolumn{2}{c}{Signal Space} \\
$\lambda_{\text{EPK}}$
& MSE $\downarrow$
& MAE $\downarrow$
& MSE $\downarrow$
& MAE $\downarrow$ \\
\midrule
0
& 0.049 & 0.167 & \textbf{0.052} & 0.117 \\
0.25
& \textbf{0.045} & \textbf{0.159}
& 0.053 & \textbf{0.110} \\
0.5
& 0.049 & 0.165 & 0.058 & 0.116 \\
0.75
& 0.047 & 0.163 & 0.053 & 0.113 \\
1.0
& 0.049 & 0.166 & 0.058 & 0.118 \\
\bottomrule
\end{tabular}
\end{table}

Table~\ref{tab:energy_weight_ablation} shows a
non-monotonic dependence on the energy coefficient.
Within this sweep, $\lambda_{\text{EPK}}=0.25$ gives the lowest
latent MSE, latent MAE, and signal MAE.
The zero-weight setting achieves slightly lower
signal MSE, 0.052 compared with 0.053.
Larger coefficients do not consistently improve
the reported metrics.

These results characterize the effect of energy-loss
weighting with EPK conditioning retained.
The zero-weight latent MSE of 0.049 and the
module-removal result of 0.061 correspond to
different configurations and should not be interpreted
as duplicate measurements of the same ablation.
The main pharmacological-mapping experiments use
$\lambda_{\text{EPK}}=0.1$ under their separately reported
protocol; this does not relabel the 0.25 setting
in the present sweep.

\subsection{Effect of the Number of Stochastic Samples}
\label{sec:res_ablationk}

We examine how the number of stochastic samples $K$
affects reconstruction performance and sampling cost.
Table~\ref{tab:lambda_ablationk} reports latent-space
and waveform-space errors for $K\in\{1,3,5\}$.

Increasing $K$ from 1 to 3 reduces all four reported
errors. In particular, signal MSE decreases from
0.056 to 0.053, while signal MAE decreases from
0.121 to 0.110.
Increasing $K$ to 5 provides further modest improvements,
with latent MSE unchanged at the reported precision.
These results indicate diminishing reconstruction gains
as the sampling budget increases.

We use $K=3$ for this reconstruction evaluation setting
to balance prediction accuracy and computational cost.
The diagnostic-score evaluation in
Section~\ref{sec:risk} separately uses $S=5$ samples
per intervention to compute the mean and variability
of diagnostic-proxy scores.
The reconstruction metrics reported here do not
directly assess uncertainty calibration.

\begin{table}[t]
\centering
\caption{
Effect of the number of stochastic samples $K$ on
reconstruction errors in latent and signal spaces.
}
\label{tab:lambda_ablationk}
\begin{tabular}{c|cc|cc}
\toprule
 & \multicolumn{2}{c|}{Latent Space}
 & \multicolumn{2}{c}{Signal Space} \\
$K$ & MSE $\downarrow$ & MAE $\downarrow$
    & MSE $\downarrow$ & MAE $\downarrow$ \\
\midrule
1 & 0.046 & 0.162 & 0.056 & 0.121 \\
3 & 0.045 & 0.159 & 0.053 & 0.110 \\
5 & 0.045 & 0.157 & 0.052 & 0.106 \\
\bottomrule
\end{tabular}
\end{table}

\subsection{Robustness under Missing-Lead Settings}
\label{sec:miss}

Table~\ref{tab:missing_leads_old_new} shows that our full model consistently outperforms the variant without EPK loss across all missing-lead settings. Improvements are observed in both latent and signal spaces, with lower MSE and MAE under every masking pattern. In particular, the performance gap becomes more pronounced under more challenging missing-lead conditions, indicating that EPK-based regularization enhances the model’s robustness to incomplete observations. By incorporating physiological priors, our method is better able to recover coherent cardiac dynamics even when input signals are partially missing.

\begin{table}[h!]
\centering
\caption{Reconstruction performance under missing-lead settings. Comparison between Ours and Ours w/o EPK loss test sets using 50 samples (lower is better).}
\label{tab:missing_leads_old_new}
\resizebox{\linewidth}{!}{
\begin{tabular}{l|cccc|cccc}
\hline
\multirow{3}{*}{Pattern} 
& \multicolumn{4}{c|}{Ours} 
& \multicolumn{4}{c}{Ours w/o EPK loss} \\
\cline{2-5} \cline{6-9}
& \multicolumn{2}{c}{Latent} 
& \multicolumn{2}{c|}{Signal}
& \multicolumn{2}{c}{Latent} 
& \multicolumn{2}{c}{Signal} \\
& MSE $\downarrow$ & MAE $\downarrow$ 
& MSE $\downarrow$ & MAE $\downarrow$
& MSE $\downarrow$ & MAE $\downarrow$ 
& MSE $\downarrow$ & MAE $\downarrow$ \\
\hline
miss-I  & 0.0451 & 0.1650 & 0.0413 & 0.1013 & 0.0485 & 0.1698 & 0.0480 & 0.1095 \\
miss-II  & 0.0454 & 0.1650 & 0.0420 & 0.1017 & 0.0484 & 0.1694 & 0.0484 & 0.1107 \\
miss-III  & 0.0449 & 0.1644 & 0.0405 & 0.1010 & 0.0476 & 0.1683 & 0.0467 & 0.1089 \\
miss-aVR  & 0.0453 & 0.1641 & 0.0399 & 0.0999 & 0.0488 & 0.1704 & 0.0487 & 0.1113 \\
miss-aVL  & 0.0426 & 0.1602 & 0.0357 & 0.0946 & 0.0489 & 0.1707 & 0.0483 & 0.1109 \\
miss-aVF  & 0.0443 & 0.1632 & 0.0381 & 0.0984 & 0.0502 & 0.1724 & 0.0496 & 0.1136 \\
miss-V1  & 0.0442 & 0.1625 & 0.0405 & 0.0999 & 0.0494 & 0.1712 & 0.0491 & 0.1112 \\
miss-V2  & 0.0458 & 0.1653 & 0.0412 & 0.1016 & 0.0487 & 0.1701 & 0.0490 & 0.1107 \\
miss-V3  & 0.0436 & 0.1611 & 0.0404 & 0.0991 & 0.0482 & 0.1695 & 0.0473 & 0.1097 \\
miss-V4  & 0.0433 & 0.1607 & 0.0395 & 0.0972 & 0.0495 & 0.1702 & 0.0496 & 0.1119 \\
miss-V5 & 0.0442 & 0.1622 & 0.0421 & 0.0995 & 0.0495 & 0.1720 & 0.0490 & 0.1127 \\
miss-V6 & 0.0445 & 0.1629 & 0.0401 & 0.0979 & 0.0510 & 0.1743 & 0.0523 & 0.1183 \\
\hline
\end{tabular}
}
\end{table}

\subsection{Counterfactual Drug-Response Simulation}

\begin{figure}[h]
    % \vspace{-.1cm}
    % \setlength{\abovecaptionskip}{-10cm}
    % \setlength{\belowcaptionskip}{-10cm}
    \centering
    \includegraphics[width=0.95\columnwidth]{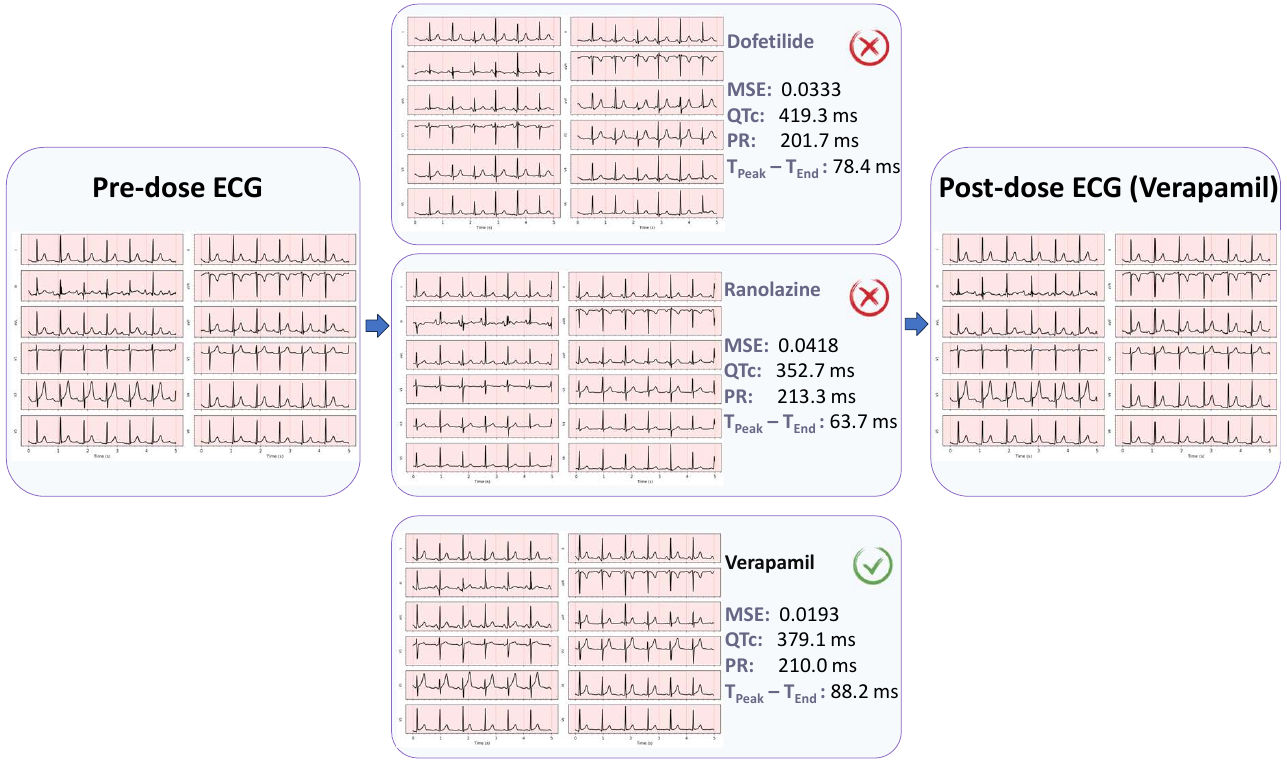}
    % \vspace{-0.6cm}
    \caption{Counterfactual drug-response simulation with the ECG World Model.
Given a normal pre-dose ECG, the model simulates post-dose outcomes under different candidate drugs, enabling direct comparison of drug-specific cardiac responses via generated ECG signals and clinical biomarkers.}
    \label{fig:v1}
    % \vspace{-.2cm}
\end{figure}

Figure~\ref{fig:v1} demonstrates the model’s ability to perform counterfactual simulation of drug responses from a single pre-dose ECG. Different candidate drugs lead to distinct post-dose ECG patterns and clinical indicators, reflecting their heterogeneous physiological effects. This enables a direct comparison of candidate interventions at the individual level, allowing the model to analyze how different treatments may alter cardiac dynamics before actual administration. This capability is valuable for personalized and risk-aware clinical analysis.

\subsection{Reconstruction under Single and Multiple Missing Leads}

\begin{figure}[h]
    % \vspace{-.1cm}
    % \setlength{\abovecaptionskip}{-10cm}
    % \setlength{\belowcaptionskip}{-10cm}
    \centering
    \includegraphics[width=0.95\columnwidth]{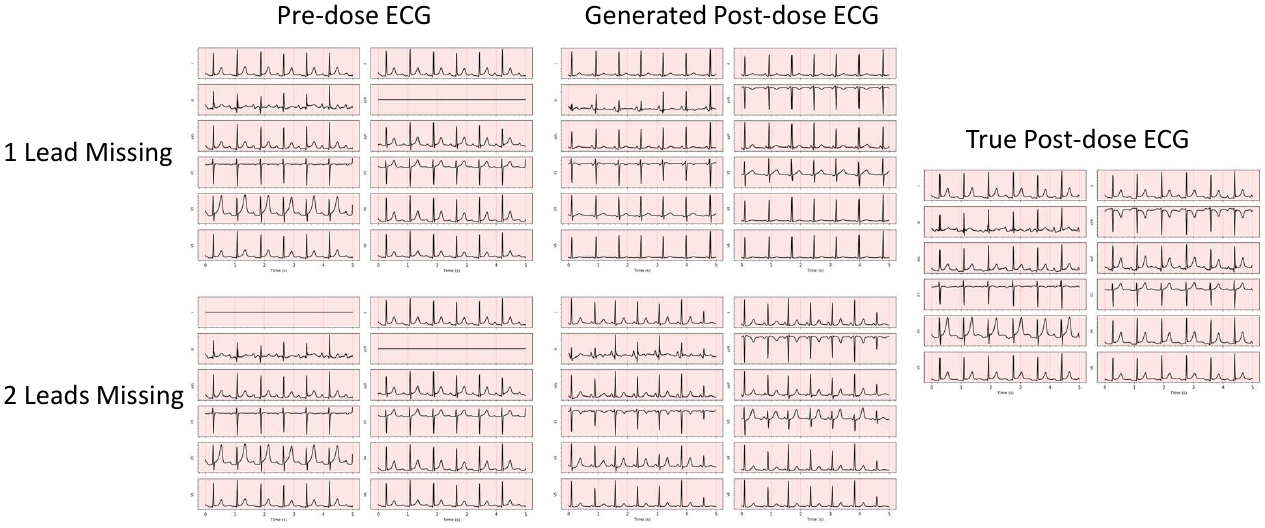}
    % \vspace{-0.6cm}
    \caption{Robustness of the predictive world model under varying degrees of data corruption. The figure demonstrates the model's ability to simultaneously forecast post-dose states and impute missing sensor data. \textbf{Left Panel (Pre-dose ECG)}: Serves as the conditioned input, with simulated sensor failures visible as flatlines. The top row illustrates a 1 Lead Missing scenario, while the bottom row demonstrates a more severe 2 Leads Missing scenario. \textbf{Middle Panel (Generated Post-dose ECG)}: Shows the model's output. Despite the corrupted pre-dose inputs, the model successfully reconstructs the missing leads while accurately predicting the morphological changes induced by the pharmacological intervention. \textbf{Right Panel (True Post-dose ECG)}: Provides the ground-truth reference. A comparison reveals that the generated waveforms—including the previously missing leads—exhibit high structural fidelity and correctly preserve inter-lead spatial correlations, indicating that the learned representation preserves useful cross-lead dependencies under the evaluated missing-lead conditions.}
    \label{fig:v12}
    % \vspace{-.2cm}
\end{figure}

To evaluate the robustness of our approach against data corruption—a prevalent issue in continuous clinical monitoring caused by electrode detachment or sensor malfunction—we investigated the model's performance under simulated single and multiple missing lead conditions. As illustrated in Figure \ref{fig:v12}, the model demonstrates exceptional fault tolerance. Even when predicting the post-dose state from pre-dose ECGs with one or two missing leads, the framework successfully reconstructs the absent signals simultaneously. Crucially, the imputed waveforms exhibit high morphological fidelity compared to the ground-truth 12-lead ECGs, preserving vital structural components (e.g., QRS complexes and T-waves) and overall rhythm dynamics without introducing perceptible artifacts or signal distortion.

This robust performance under extreme data degradation scenarios serves as compelling evidence for the efficacy of our proposed world model paradigm. Conventional direct mapping networks typically suffer severe performance degradation or catastrophic failure when confronted with incomplete input modalities, as they heavily rely on complete point-to-point correspondences. In contrast, our world model does not merely perform surface-level signal interpolation. Instead, it has internalized the intrinsic spatio-temporal dynamics and structural dependencies of the cardiovascular system's electrical activity.

By establishing a robust latent representation of this physiological "world," the model treats the missing leads as partial observations. It effectively leverages the spatial correlations inherent in the remaining functional leads to accurately infer the complete, unobserved physiological state. The ability to synthesize coherent, high-fidelity 12-lead ECGs from severely corrupted inputs—whether a single lead or multiple leads are compromised—underscores that the model has learned representations that support coherent reconstruction under the evaluated perturbations.  This intrinsic understanding provides empirical evidence of robustness within the evaluated settings. %and continuous, robust forecasting even in highly suboptimal clinical scenarios, thereby highlighting the immense clinical utility of world models in medical time-series analysis.

\subsection{Reconstruction under Single-Lead Missing in Abnormal Patients}

\begin{figure}[h]
    % \vspace{-.1cm}
    % \setlength{\abovecaptionskip}{-10cm}
    % \setlength{\belowcaptionskip}{-10cm}
    \centering
    \includegraphics[width=0.95\columnwidth]{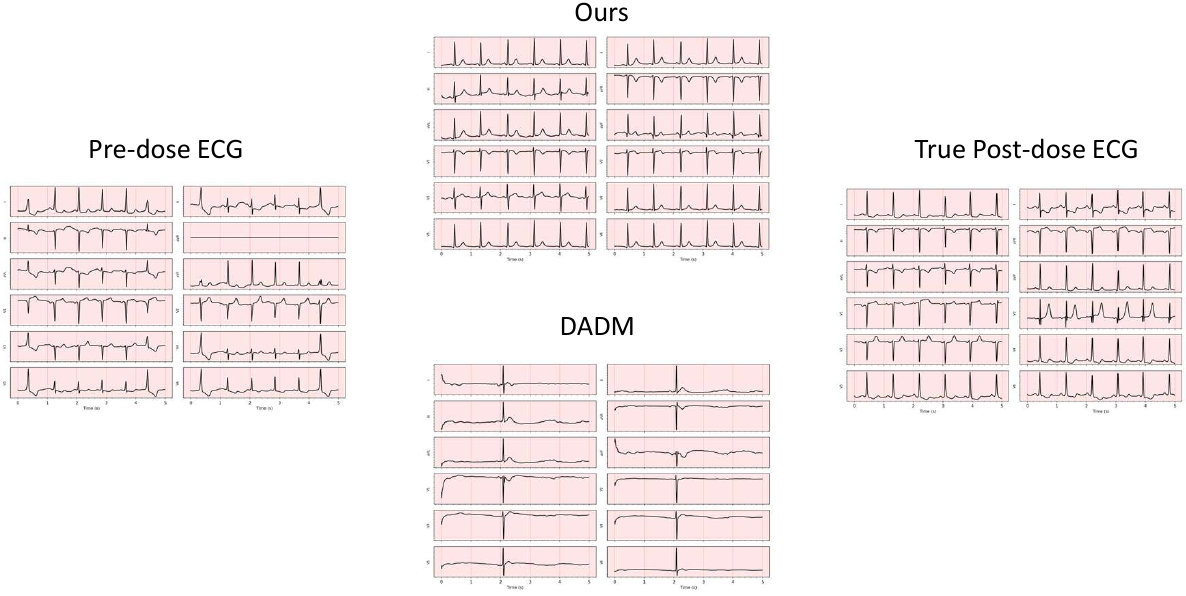}
    % \vspace{-0.6cm}
    \caption{Reconstruction under single-lead missing in an abnormal exemplar patient. Given pre-dose ECGs with one lead removed, our method reconstructs post-dose waveforms that closely match the ground truth, effectively preserving pathological characteristics. In contrast, DADM produces oversmoothed signals and fails to recover clinically relevant patterns, leading to noticeable deviations from the true post-dose ECG.}
    \label{fig:v22}
    % \vspace{-.2cm}
\end{figure}

Under the single-lead missing setting, the reconstruction task becomes particularly challenging for abnormal patients, where critical diagnostic information may be distributed across leads. As shown in Figure \ref{fig:v22}, our method is able to accurately infer the missing information and recover post-dose ECG waveforms that remain highly consistent with the ground truth. Notably, key pathological patterns are well preserved, indicating that the model effectively captures cross-lead dependencies and underlying physiological dynamics.

For instance, in a representative case with complex cardiovascular conditions—subendocardial infarction (initial episode of care), acute on chronic systolic heart failure, and coronary artery dissection—the patient had a pre-dose potassium level of 3.9 mmol/L and was administered Furosemide (Lasix) combined with Heparin Sodium. Despite the presence of multi-factorial cardiac abnormalities and pharmacological intervention, the reconstructed ECG successfully preserves clinically relevant waveform characteristics, further demonstrating the robustness of our approach under challenging pathological scenarios.

In contrast, DADM struggles in this setting, producing oversmoothed waveforms with diminished abnormal features. This suggests that it fails to leverage inter-lead relationships when partial observations are provided, resulting in degraded reconstruction quality. Overall, these results highlight the advantage of our approach in handling incomplete inputs and maintaining clinically meaningful signal structures under challenging abnormal conditions.

\subsection{Stability and Graceful Degradation}
\label{b9}

\begin{figure}[h]
    % \vspace{-.1cm}
    % \setlength{\abovecaptionskip}{-10cm}
    % \setlength{\belowcaptionskip}{-10cm}
    \centering
    \includegraphics[width=0.5\columnwidth]{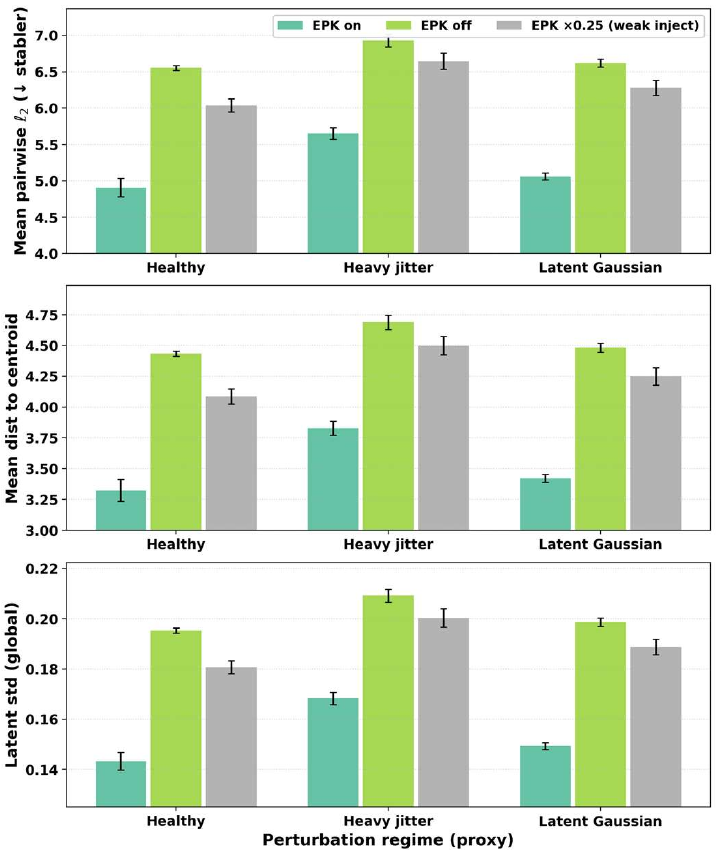}
    % \vspace{-0.6cm}
    \caption{Empirical stability of the EPK mechanism under increasing prior mismatch. We compare full prior injection ($\text{EPK on}$), weak prior ($\text{EPK} \times 0.25$), and no prior ($\text{EPK off}$) across three perturbation regimes: Healthy, Heavy Jitter, and Latent Gaussian. Stability is quantified via three complementary metrics: Mean Pairwise $\ell_2$ Distance, Mean Distance to Centroid, and global Latent Standard Deviation (lower is better for all). Across all regimes, enabling EPK significantly reduces generation variability and latent dispersion. Notably, under severe prior mismatch, EPK-enabled models exhibit graceful degradation—stability metrics degrade smoothly and remain bounded without catastrophic failure, consistently outperforming the baseline without prior injection.}
    \label{fig:v32}
    % \vspace{-.2cm}
\end{figure}

As illustrated in Figure~\ref{fig:v32}, our empirical evaluation reveals a consistent ordering across all three perturbation regimes (Healthy, Heavy Jitter, and Latent Gaussian): $\text{EPK on} < \text{EPK} \times 0.25 < \text{EPK off}$. This trend is strictly preserved across all evaluated metrics, Mean Pairwise $\ell_2$ Distance, Mean Distance to Centroid, and global Latent Standard Deviation. The simultaneous reduction in inter-sample variability and latent dispersion confirms that the EPK mechanism effectively enhances overall generation stability.Crucially, as the system is subjected to increasing prior mismatch, moving from the well-aligned Healthy regime to unstructured Latent Gaussian noise, the degradation in stability is smooth and bounded. While all configurations experience moderate performance drops due to the growing divergence between the ODE prior and the true data dynamics, the EPK-enabled models do not exhibit abrupt spikes or catastrophic collapse. Furthermore, the intermediate performance of the weak prior ($\text{EPK} \times 0.25$) indicates that the influence of the prior is continuous and tunable via scaling, rather than acting as a binary constraint.Together, these observations provide strong evidence that our model exhibits graceful degradation. Rather than failing catastrophically when the prior becomes unreliable, the system smoothly transitions toward data-driven behavior. It maintains a strict stability advantage over the unguided baseline ($\text{EPK off}$), demonstrating robust performance even under severe prior misalignment.

\subsection{Clinical Risk Formulation and Label Aggregation}
\label{scr}

To quantitatively assess diagnostic changes in simulated
pharmacological responses, we map generated ECG waveforms
to a scalar diagnostic risk proxy using a frozen multi-label
ECG-FM model, denoted as $f_{\mathrm{risk}}$.

\paragraph{Diagnostic Model and 17-Dimensional Target Labels}
The diagnostic model is formulated around the
MIMIC-IV-ECG machine reads task as a multi-label
classification problem with 17 binary targets.
Each target indicates the presence of a particular
interpretation in the machine-generated diagnostic report.
The categories are: Poor data quality, Sinus rhythm, PVC,
Tachycardia, Ventricular tachycardia, SVT with aberrancy,
Atrial fibrillation, Atrial flutter, Bradycardia,
Accessory pathway conduction, AV block, 1st degree AV block,
Bifascicular block, RBBB, LBBB, Myocardial infarction,
and Electronic pacemaker.

\paragraph{Aggregation into a Scalar Risk Proxy}
For an ECG waveform $x$, the model $f_{\mathrm{risk}}$
outputs 17 logits. Let $\ell_j(x)$ denote the logit for
label $j$. Applying a sigmoid to each logit gives the
corresponding diagnostic probability:
\[
p_j(x)
=
\operatorname{sigmoid}\!\left(\ell_j(x)\right),
\qquad j=1,\dots,17.
\]
The implemented aggregation is the arithmetic mean
of these probabilities:
\begin{equation}
R(x)
=
\frac{1}{17}\sum_{j=1}^{17}p_j(x).
\end{equation}
The same frozen model and aggregation rule are applied
to generated, observed post-dose, and pre-dose ECGs.

For each candidate intervention $a$, we generate $S=5$
next-state samples using standard DDPM with EPK conditioning:
\[
\hat{z}_{k+1}^{(s)}
\sim p_{\mathrm{WM}}(\cdot\mid z_k,a),
\qquad s=1,\dots,S.
\]
Scoring each decoded waveform yields
\begin{equation}
R^{(s)}(a)
=
R\!\left(
f_{\mathrm{dec}}(\hat{z}_{k+1}^{(s)})
\right).
\end{equation}
We summarize the scores using their empirical mean
and variance:
\begin{equation}
\mu_R(a)
=
\frac{1}{S}\sum_{s=1}^{S}R^{(s)}(a),
\qquad
\sigma_R^2(a)
=
\frac{1}{S-1}\sum_{s=1}^{S}
\left(R^{(s)}(a)-\mu_R(a)\right)^2.
\end{equation}
These quantities characterize the mean diagnostic score
under the world model and its variability across
stochastic simulations.

For the recorded medication $a_{\mathrm{true}}$,
predicted and observed diagnostic-score changes use
the same pre-dose reference:
\begin{align}
\Delta_{\mathrm{wm}}
&=
\mu_R(a_{\mathrm{true}})-R(x_{\mathrm{pre}}),
\\
\Delta_{\mathrm{real}}
&=
R(x_{\mathrm{post}})-R(x_{\mathrm{pre}}).
\end{align}
The risk-aversion coefficient $\lambda$ affects candidate
selection through $\mu_R(a)+\lambda\sigma_R(a)$,
but does not enter these diagnostic-change calculations.
Comparing $\Delta_{\mathrm{wm}}$ with $\Delta_{\mathrm{real}}$
assesses diagnostic-change fidelity under a shared backend.
Because the label set includes heterogeneous findings,
the aggregate is treated as an evaluation proxy rather
than a calibrated measure of overall clinical severity.

\section{Derivation of the ECG World Model}

This appendix provides a detailed derivation of the proposed ECG world model. We first introduce the notation and assumptions, then derive the diffusion process and its reverse parameterization, and finally present the physiological prior, training objective, and inference procedure.

\subsection{Notations and Basic Assumptions}

\subsubsection{Time-Scale Separation}
We distinguish between two time scales:

\begin{itemize}[leftmargin=10pt]
    \item \textbf{MDP timestep}: indexed by $k = 0,1,\dots,L$. At step $k$, the true latent cardiac state is denoted by $z_k \in \mathbb{R}^d$, the model prediction is $\hat{z}_k$, and the applied medical action is $a_k \in \mathcal{A}$. The environment evolves according to
    \begin{equation}
        z_{k+1} = \mathcal{T}(z_k, a_k) + \eta_k, 
        \qquad 
        \eta_k \sim \mathcal{N}(0, \sigma_\eta^2 I_d),
        \label{eq:true_dynamics}
    \end{equation}
    where $I_d$ is the $d \times d$ identity matrix and $\eta_k$ is independent of the past conditioned on $(z_k, a_k)$.

    \item \textbf{Diffusion timestep}: indexed by $\tau = 0,1,\dots,N$, where $\tau=0$ denotes the clean latent sample and $\tau=N$ denotes pure noise. For a fixed MDP step $k$, the diffusion model generates $z_{k+1}$ by denoising from Gaussian noise, conditioned on the current estimate $\hat{z}_k$ and the action $a_k$.
\end{itemize}

\subsubsection{VAE and Latent-Space Assumptions}
We assume an encoder $f_{\mathrm{enc}} : \mathbb{R}^{C \times T_{\mathrm{sig}}} \to \mathbb{R}^d$ and a decoder $f_{\mathrm{dec}} : \mathbb{R}^d \to \mathbb{R}^{C \times T_{\mathrm{sig}}}$ such that
$f_{\mathrm{dec}}(f_{\mathrm{enc}}(x)) \approx x$.
For analysis, we assume that the decoder is locally bi-Lipschitz on a compact subset of the physiological manifold $\mathcal{M} \subset \mathbb{R}^d$:
\begin{equation}
    \exists\, 0 < C_1 \le C_2,\quad \forall z,z' \in \mathcal{M},\qquad
    C_1 \|z-z'\|_2 \le \|f_{\mathrm{dec}}(z)-f_{\mathrm{dec}}(z')\|_2 \le C_2 \|z-z'\|_2.
    \label{eq:bi_lipschitz}
\end{equation}
This assumption explicitly connects the bounded multi-step latent-space errors to waveform-space structural divergence in our analysis.

\subsection{Diffusion Model Basics}
For simplicity, we omit the explicit condition $(\hat{z}_k, a_k)$ in this subsection, although all distributions are conditioned on them.

\subsubsection{Forward Diffusion Process}
Let $\beta_1,\beta_2,\dots,\beta_N \in (0,1)$ be a variance schedule, and define
\[
\alpha_\tau = 1-\beta_\tau,\qquad \bar{\alpha}_\tau = \prod_{i=1}^{\tau} \alpha_i.
\]
The forward diffusion process is the Markov chain
\begin{equation}
    q(z_{k+1}^{(\tau)} \mid z_{k+1}^{(\tau-1)})
    =
    \mathcal{N}\!\left(
        z_{k+1}^{(\tau)};\,
        \sqrt{\alpha_\tau}\, z_{k+1}^{(\tau-1)},
        \beta_\tau I_d
    \right),
    \qquad \tau=1,\dots,N,
\end{equation}
with $z_{k+1}^{(0)} = z_{k+1}$.
By iterating the recursion, we obtain the standard closed form
\begin{equation}
    q(z_{k+1}^{(\tau)} \mid z_{k+1}^{(0)})
    =
    \mathcal{N}\!\left(
        z_{k+1}^{(\tau)};\,
        \sqrt{\bar{\alpha}_\tau}\, z_{k+1}^{(0)},
        (1-\bar{\alpha}_\tau) I_d
    \right).
    \label{eq:forward_marginal}
\end{equation}

\paragraph{Derivation}
The result follows by induction. For $\tau=1$, the claim is immediate from the definition. Assume it holds for $\tau-1$. Then
\begin{align*}
    z_{k+1}^{(\tau)}
    &= \sqrt{\alpha_\tau}\, z_{k+1}^{(\tau-1)} + \sqrt{\beta_\tau}\,\epsilon_\tau \\
    &= \sqrt{\alpha_\tau}\left(\sqrt{\bar{\alpha}_{\tau-1}}\, z_{k+1}^{(0)} + \sqrt{1-\bar{\alpha}_{\tau-1}}\, \epsilon'\right)
       + \sqrt{\beta_\tau}\,\epsilon_\tau \\
    &= \sqrt{\bar{\alpha}_\tau}\, z_{k+1}^{(0)}
       + \sqrt{\alpha_\tau(1-\bar{\alpha}_{\tau-1})}\,\epsilon'
       + \sqrt{\beta_\tau}\,\epsilon_\tau,
\end{align*}
where $\epsilon',\epsilon_\tau \sim \mathcal{N}(0,I_d)$ are independent.
Since the last two terms are independent Gaussians, their sum is Gaussian with covariance
\[
\alpha_\tau(1-\bar{\alpha}_{\tau-1}) + \beta_\tau
= 1 - \bar{\alpha}_\tau,
\]
which proves Equation~\ref{eq:forward_marginal}.

\subsubsection{Reverse Process and Parameterization}
The exact reverse posterior has the form
\begin{equation}
    q\!\left(z_{k+1}^{(\tau-1)} \mid z_{k+1}^{(\tau)}, z_{k+1}^{(0)}\right)
    =
    \mathcal{N}\!\left(
        z_{k+1}^{(\tau-1)};\,
        \tilde{\mu}_\tau(z_{k+1}^{(\tau)}, z_{k+1}^{(0)}),
        \tilde{\beta}_\tau I_d
    \right),
\end{equation}
where
\begin{equation}
    \tilde{\beta}_\tau
    =
    \frac{1-\bar{\alpha}_{\tau-1}}{1-\bar{\alpha}_\tau}\,\beta_\tau,
\end{equation}
and
\begin{equation}
    \tilde{\mu}_\tau(z_{k+1}^{(\tau)}, z_{k+1}^{(0)})
    =
    \frac{\sqrt{\bar{\alpha}_{\tau-1}}\beta_\tau}{1-\bar{\alpha}_\tau}\, z_{k+1}^{(0)}
    +
    \frac{\sqrt{\alpha_\tau}(1-\bar{\alpha}_{\tau-1})}{1-\bar{\alpha}_\tau}\, z_{k+1}^{(\tau)}.
    \label{eq:posterior_mean}
\end{equation}

We parameterize the learned reverse process by predicting the clean latent sample. Let
\[
\hat{z}_\theta
=
\hat{z}_\theta\!\left(z_{k+1}^{(\tau)}, \tau, \hat{z}_k, a_k\right)
\]
denote the network prediction of $z_{k+1}^{(0)}$. Then the learned reverse transition is
\begin{equation}
    p_\theta\!\left(z_{k+1}^{(\tau-1)} \mid z_{k+1}^{(\tau)}, \hat{z}_k, a_k\right)
    =
    \mathcal{N}\!\left(
        z_{k+1}^{(\tau-1)};\,
        \mu_\theta\!\left(z_{k+1}^{(\tau)}, \tau, \hat{z}_k, a_k\right),
        \tilde{\beta}_\tau I_d
    \right),
\end{equation}
with
\begin{equation}
    \mu_\theta\!\left(z_{k+1}^{(\tau)}, \tau, \hat{z}_k, a_k\right)
    =
    \tilde{\mu}_\tau\!\left(z_{k+1}^{(\tau)}, \hat{z}_\theta\!\left(z_{k+1}^{(\tau)}, \tau, \hat{z}_k, a_k\right)\right).
    \label{eq:reverse_mean}
\end{equation}
Equivalently, if the network predicts noise $\epsilon_\theta$, then the clean estimate is
\begin{equation}
    \hat{z}_\theta
    =
    \frac{1}{\sqrt{\bar{\alpha}_\tau}}
    \left(
        z_{k+1}^{(\tau)} - \sqrt{1-\bar{\alpha}_\tau}\,\epsilon_\theta(z_{k+1}^{(\tau)}, \tau, \hat{z}_k, a_k)
    \right),
\end{equation}
and substituting this expression into Equation~\ref{eq:reverse_mean} yields the standard DDPM reverse parameterization.

\subsection{Construction of the External Physiological Prior (EPK)}

\paragraph{ECG ODE}
The ECG waveform $y(\theta)$, viewed as a function of the
cardiac phase angle $\theta$, is governed by
\begin{equation}
    \frac{dy}{d\theta}
    = -\sum_{i \in \{P,Q,R,S,T\}}
    \alpha_i \Delta\theta_i
    \exp\!\left(-\frac{\Delta\theta_i^2}{2b_i^2}\right)
    - (y-y_0),
    \label{eq:mcsharry}
\end{equation}
where $\Delta\theta_i=(\theta-\theta_i)\bmod 2\pi$, and
$\Theta=\{\alpha_i,b_i,\theta_i,y_0\}_{i\in\{P,Q,R,S,T\}}$
denotes the physiological parameters controlling the
amplitude, width, center position, and baseline of the
wave components.

\paragraph{Parameter Modulation under Interventions}
Given a pharmacological action $a\in\mathcal{A}$, we define
a modulation map
\begin{equation}
    \Theta_a=\mathcal{M}(a,\Theta),
\end{equation}
where $\mathcal{M}$ adjusts the ODE parameters according
to the expected physiological effect of the intervention.
In our implementation, $\mathcal{M}$ is a rule-based
heuristic informed by pharmacological literature.
It scales selected baseline parameters, such as the
wave widths $\{b_i\}$, to construct regimen-dependent
physiological templates.

\paragraph{Estimating the Initial Phase from the Current State}
Given the estimated latent state $\hat{z}_k$ at MDP step
$k$, we decode it into an ECG waveform:
\begin{equation}
    \hat{x}_k=f_{\mathrm{dec}}(\hat{z}_k).
\end{equation}
A deterministic detector $h$ extracts an initial phase
anchor from the waveform:
\begin{equation}
    \theta_{\mathrm{start}}
    =h(\hat{x}_k)\in[0,2\pi).
\end{equation}
In practice, $h$ uses Pan--Tompkins R-peak localization
to establish the phase anchor.

\paragraph{ODE Integration and Projection}
Given a prediction horizon $\Delta\theta_{\mathrm{pred}}$,
let $y^{(a)}(\theta)$ denote the physiological waveform
obtained by integrating Equation~\ref{eq:mcsharry} under
the modulated parameters $\Theta_a$.
Here, $y$ denotes the waveform-level amplitude, distinct
from the latent state $z$ used by the VAE and diffusion
model. Starting from $\theta_{\mathrm{start}}$, the
trajectory evolves along the phase coordinate as
\begin{equation}
\begin{aligned}
    y^{(a)}(\theta)
    &=
    y(\theta_{\mathrm{start}})
    +
    \int_{\theta_{\mathrm{start}}}^{\theta}
    \mathcal{F}\!\left(y^{(a)}(u),u;\Theta_a\right)\,du,
    \\
    &\qquad
    \theta\in
    [\theta_{\mathrm{start}},
    \theta_{\mathrm{start}}+\Delta\theta_{\mathrm{pred}}],
\end{aligned}
\end{equation}
where $\mathcal{F}$ denotes the right-hand side of
Equation~\ref{eq:mcsharry}.

We numerically integrate the ODE and sample the resulting
trajectory at $L$ uniformly spaced phase points:
\begin{equation}
    e_{\mathrm{EPK}}
    =
    \left\{
    y^{(a)}\!\left(
    \theta_{\mathrm{start}}
    +\frac{l}{L}\Delta\theta_{\mathrm{pred}}
    \right)
    \right\}_{l=1}^{L}
    \in\mathbb{R}^{L}.
\end{equation}
The physiological sequence is mapped into the latent
space through the projection module $\Pi_\phi$:
\begin{equation}
    z_{\mathrm{EPK}}(\hat{z}_k,a_k)
    =
    \Pi_\phi\!\big(
    e_{\mathrm{EPK}}(\hat{z}_k,a_k)
    \big)
    \in\mathbb{R}^{d}.
    \label{eq:epk_projection}
\end{equation}
The projection consists of temporal resampling to the
latent sequence length followed by a learnable
$1\times1$ convolution.
Its parameters $\phi$ are jointly optimized with the
diffusion model parameters $\theta$, while the
pre-trained VAE remains frozen.
The projected anchor supplies the target for the
latent physiological energy term; EPK conditioning
is provided through a separate embedding pathway.

\paragraph{Remark}
For fixed model parameters, $z_{\mathrm{EPK}}$ depends
on the current estimated state $\hat{z}_k$ and action
$a_k$, but not on the diffusion timestep $\tau$.
It therefore provides a common physiological anchor
for the next-step clean-latent predictions across
diffusion timesteps. The decoding used for phase
initialization is distinct from evaluating the energy
penalty, which operates directly in latent space.

\subsection{Derivation of the Training Loss}

\subsubsection{Standard Diffusion Loss}
Let $z_{k+1}^{(0)}$ denote the clean next-step latent sample.
We sample a diffusion timestep
$\tau\sim\mathrm{Uniform}\{1,\ldots,N\}$ and construct
\[
z_{k+1}^{(\tau)}
=
\sqrt{\bar{\alpha}_\tau}\,z_{k+1}^{(0)}
+
\sqrt{1-\bar{\alpha}_\tau}\,\epsilon,
\qquad
\epsilon\sim\mathcal{N}(0,I_d).
\]
The denoiser predicts the injected noise:
\[
\hat{\epsilon}
=
\epsilon_\theta\!\left(
z_{k+1}^{(\tau)},\tau,\hat{z}_k,a_k
\right),
\]
where additional conditioning inputs, including the EPK
embedding, are omitted for brevity.
The implemented diffusion loss is the unweighted
noise-prediction objective:
\begin{equation}
    \mathcal{L}_{\mathrm{data}}(\theta)
    =
    \mathbb{E}
    \left[
        \left\|\hat{\epsilon}-\epsilon\right\|_2^2
    \right].
    \label{eq:loss_data}
\end{equation}
The expectation is taken over training transitions,
diffusion timesteps, and Gaussian noise.

\paragraph{Interpretation.}
The corresponding clean-latent estimate is derived from
the predicted noise:
\[
\hat{z}_0
=
\hat{z}_\theta\!\left(
z_{k+1}^{(\tau)},\tau,\hat{z}_k,a_k
\right)
=
\frac{
z_{k+1}^{(\tau)}
-
\sqrt{1-\bar{\alpha}_\tau}\,\hat{\epsilon}
}{
\sqrt{\bar{\alpha}_\tau}
}.
\]
Here, $\hat{z}_\theta$ denotes the derived clean estimate,
not a separate prediction network.
This estimate is used for physiological regularization,
while the data loss in Equation~\ref{eq:loss_data}
is evaluated in noise space.

\subsubsection{Physiological Regularization Term}
We regularize the predicted clean sample rather than the
noisy intermediate state. The forward process introduces
Gaussian noise for diffusion training; applying the
physiological penalty to $\hat{z}_0$ instead of
$z_{k+1}^{(\tau)}$ targets the denoised representation.

We define the physiological residual as
\begin{equation}
    \mathcal{R}_{\mathrm{phys}}(\theta,\phi)
    =
    \left\|
        \hat{z}_0
        -
        z_{\mathrm{EPK}}(\hat{z}_k,a_k)
    \right\|_2^2.
\end{equation}
The anchor is obtained through the EPK projection:
\[
z_{\mathrm{EPK}}(\hat{z}_k,a_k)
=
\Pi_\phi\!\left(
e_{\mathrm{EPK}}(\hat{z}_k,a_k)
\right).
\]

\paragraph{Connection to the EPK Energy.}
This squared residual is the latent energy introduced
in the main text:
$
E_\phi(\hat{z}_0)
=
\mathcal{R}_{\mathrm{phys}}(\theta,\phi).
$
It measures deviation from the intervention-conditioned
physiological anchor directly in latent space, without
waveform decoding to evaluate the penalty.

The physiological loss uses a fixed coefficient:
\begin{equation}
    \mathcal{L}_{\mathrm{phys}}(\theta,\phi)
    =
    \lambda_{\text{EPK}}\,
    \mathbb{E}
    \left[
        \mathcal{R}_{\mathrm{phys}}(\theta,\phi)
    \right].
    \label{eq:loss_phys}
\end{equation}
There is no additional timestep-dependent weighting.
The main pharmacological-mapping experiments use
$\lambda_{\text{EPK}}=0.1$, while the energy-weight
ablation uses its separately reported coefficients.

\paragraph{Training convention}
During training, we use teacher forcing by setting
$\hat{z}_k=z_k$.
The physiological anchor is constructed from this
current state and the recorded action $a_k$.
For a fixed transition and fixed projection parameters,
the anchor does not depend on the diffusion timestep
$\tau$, although the predicted clean latent does.

\subsubsection{Joint Objective}
The final training objective combines the noise-prediction
loss and physiological regularization:
\begin{equation}
\begin{aligned}
    \mathcal{L}_{\mathrm{ECGWM}}(\theta,\phi)
    &=
    \mathcal{L}_{\mathrm{data}}(\theta)
    +
    \mathcal{L}_{\mathrm{phys}}(\theta,\phi)
    \\
    &=
    \mathbb{E}
    \left[
        \left\|\hat{\epsilon}-\epsilon\right\|_2^2
        +
        \lambda_{\text{EPK}}
        \left\|
            \hat{z}_0
            -
            z_{\mathrm{EPK}}(\hat{z}_k,a_k)
        \right\|_2^2
    \right].
\end{aligned}
    \label{eq:loss_joint}
\end{equation}
The diffusion parameters $\theta$ and projection
parameters $\phi$ are jointly optimized, while the
pre-trained VAE remains frozen.

\paragraph{Summary}
Equation~\ref{eq:loss_joint} combines conditional noise
prediction with physiological anchoring of the derived
clean latent. The first term learns denoising from data,
while the second encourages agreement with the projected
intervention-conditioned prior.

\subsection{Inference-Time Behavior under Physiological Regularization}

Although physiological regularization embeds domain knowledge during training, the diffusion model remains stochastic at inference time.

In this work, we do not perform explicit energy-based resampling during inference. Instead, we follow the standard diffusion sampling procedure, where the next-step latent state is generated by iteratively denoising from Gaussian noise using the learned reverse process:
\begin{equation}
    \hat{z}_{k+1} \sim p_\theta(\cdot \mid \hat{z}_k, a_k).
\end{equation}

\paragraph{Implicit knowledge injection}
The physiological constraint introduced by the EPK module is incorporated implicitly through the energy-regularized training objective. Specifically, the learned model $p_\theta$ is biased toward generating samples that are close to the physiological anchor $z_{\mathrm{EPK}}(\hat{z}_k, a_k)$ in latent space.

\paragraph{Effective generative distribution}
As a result, the effective generative behavior can be interpreted as approximately sampling from a distribution that favors physiologically consistent states:
\begin{equation}
    p_{\theta,\phi}(z_{k+1} \mid \hat{z}_k, a_k)
    \approx
    p_\theta(z_{k+1} \mid \hat{z}_k, a_k),
\end{equation}
where the influence of the physiological prior is already absorbed into the learned parameters $\theta$ during training.

\paragraph{Closed-loop generative operator}
We denote the resulting generative process by
\begin{equation}
    \hat{z}_{k+1} \sim \mathcal{G}_\theta(\cdot \mid \hat{z}_k, a_k),
\end{equation}
where $\mathcal{G}_\theta$ represents the diffusion model trained with physiological regularization.

\paragraph{Remark}
Although explicit energy-based reweighting or resampling could further enforce physiological constraints at inference time, we find that incorporating the EPK prior during training is sufficient to produce stable and physiologically plausible trajectories in practice.

\subsection{Multi-Step Prediction Error Analysis}

\subsubsection{Error Definition and Recurrence}
Define the prediction error at step $k$ as $\delta_k = \hat{z}_k - z_k$.
Using Equation~\ref{eq:true_dynamics} and the learned generative operator, we obtain
\begin{equation}
    \delta_{k+1}
    =
    \mathcal{G}_{\theta}(\hat{z}_k, a_k)
    -
    \mathcal{T}(z_k, a_k)
    -
    \eta_k,
    \label{eq:error_step}
\end{equation}
Taking conditional expectation and using $\mathbb{E}[\eta_k]=0$ and $\mathbb{E}[\|\eta_k\|_2^2]=\sigma_\eta^2 d$, we get
\begin{equation}
    \mathbb{E}\!\left[\|\delta_{k+1}\|_2^2\right]
    =
    \mathbb{E}\!\left[\big\|\mathcal{G}_{\theta}(\hat{z}_k, a_k)-\mathcal{T}(z_k, a_k)\big\|_2^2\right]
    + \sigma_\eta^2 d.
\end{equation}

\subsubsection{Contraction Assumption}
\begin{assumption}[Contractive closed-loop dynamics]
\label{assum:contraction}
There exist $\kappa \in (0,1)$ and $\Delta_{\mathrm{sys}} \ge 0$ such that for any $z,\hat{z}\in\mathbb{R}^d$ and any action $a\in\mathcal{A}$,
\begin{equation}
    \mathbb{E}\!\left[
        \big\|\mathcal{G}_{\theta}(\hat{z}, a)-\mathcal{T}(z, a)\big\|_2^2
    \right]
    \le
    \kappa \|\hat{z}-z\|_2^2 + \Delta_{\mathrm{sys}}.
\end{equation}
\end{assumption}

\paragraph{Scope of the contraction assumption.}
The condition $\kappa<1$ is an assumption of the error
analysis, rather than a property established by the
training objective.
Our short-chain rollout experiments show limited additional
terminal error under the evaluated conditions, but do not
verify global contraction or guarantee long-horizon stability.

\paragraph{Extension to Waveform Space} 
Crucially, utilizing the bi-Lipschitz assumption established in Equation~\ref{eq:bi_lipschitz}, the expected error in the physical continuous waveform space is strictly bounded by the latent space error:
\begin{equation}
    \mathbb{E}\!\left[\big\|\hat{x}_{k+1} - x_{k+1}\big\|_2^2\right] \le C_2^2 \, \mathbb{E}\!\left[\|\delta_{k+1}\|_2^2\right].
\end{equation}
This formalizes that an EPK-induced contractive latent dynamic ($\kappa < 1$) strictly prevents compounding divergence in the macroscopic physiological signal space.

\subsection{Algorithm Pseudocodes}

Both algorithms retain EPK conditioning.
We include $e_{\mathrm{EPK}}$ as a conditioning argument,
with its embedding transformation and other clinical
conditioning inputs omitted for brevity.

\begin{algorithm}[H]
\caption{Training Algorithm for ECG World Model}
\begin{algorithmic}[1]
\REQUIRE Dataset $\mathcal{D}$, frozen VAE
$(f_{\mathrm{enc}},f_{\mathrm{dec}})$, EPK module,
coefficient $\lambda_{\text{EPK}}$,
diffusion schedule $\{\beta_\tau\}_{\tau=1}^{N}$
\STATE Initialize diffusion parameters $\theta$
and EPK projection parameters $\phi$
\REPEAT
    \STATE Sample an MDP transition
    $(z_k,z_{k+1}^{(0)})$ and action $a_k$
    from $\mathcal{D}$
    \STATE Set $\hat{z}_k \leftarrow z_k$
    \COMMENT{Teacher forcing}
    \STATE Construct the physiological sequence:
    $e_{\mathrm{EPK}}
    \leftarrow e_{\mathrm{EPK}}(\hat{z}_k,a_k)$
    \STATE Sample
    $\tau\sim\mathrm{Uniform}\{1,\dots,N\}$
    and $\epsilon\sim\mathcal{N}(0,I_d)$
    \STATE Construct the noisy latent:
    \[
    z_{k+1}^{(\tau)}
    =
    \sqrt{\bar{\alpha}_\tau}\,z_{k+1}^{(0)}
    +
    \sqrt{1-\bar{\alpha}_\tau}\,\epsilon
    \]
    \STATE Predict the injected noise:
    \[
    \hat{\epsilon}
    =
    \epsilon_\theta\!\left(
    z_{k+1}^{(\tau)},\tau,\hat{z}_k,a_k,
    e_{\mathrm{EPK}}
    \right)
    \]
    \STATE Compute the predicted clean latent:
    \[
    \hat{z}_0
    =
    \frac{
    z_{k+1}^{(\tau)}
    -
    \sqrt{1-\bar{\alpha}_\tau}\,\hat{\epsilon}
    }{
    \sqrt{\bar{\alpha}_\tau}
    }
    \]
    \STATE Compute the physiological anchor:
    $z_{\mathrm{EPK}}
    \leftarrow \Pi_\phi(e_{\mathrm{EPK}})$
    \STATE Compute the joint loss:
    \[
    \mathcal{L}
    =
    \|\hat{\epsilon}-\epsilon\|_2^2
    +
    \lambda_{\text{EPK}}
    \|\hat{z}_0-z_{\mathrm{EPK}}\|_2^2
    \]
    \STATE Update $\theta$ and $\phi$ to minimize
    $\mathcal{L}$, keeping the VAE frozen
\UNTIL{the training budget is reached}
\end{algorithmic}
\end{algorithm}

\begin{algorithm}[H]
\caption{Inference Algorithm for ECG World Model}
\begin{algorithmic}[1]
\REQUIRE Initial state $z_0$,
action sequence $a_0,\dots,a_{L-1}$,
trained model parameters $\theta$,
frozen VAE $(f_{\mathrm{enc}},f_{\mathrm{dec}})$,
EPK module, diffusion schedule
$\{\beta_\tau\}_{\tau=1}^{N}$
\STATE $\hat{z}_0 \leftarrow z_0$
\FOR{$k=0$ to $L-1$}
    \STATE Construct the physiological sequence:
    $e_{\mathrm{EPK}}
    \leftarrow e_{\mathrm{EPK}}(\hat{z}_k,a_k)$
    \STATE Sample
    $z_{k+1}^{(N)}\sim\mathcal{N}(0,I_d)$
    \FOR{$\tau=N$ down to $1$}
        \STATE Sample using the standard DDPM
        reverse transition:
        \[
        z_{k+1}^{(\tau-1)}
        \sim
        p_\theta\!\left(
        \cdot\mid
        z_{k+1}^{(\tau)},\tau,\hat{z}_k,a_k,
        e_{\mathrm{EPK}}
        \right)
        \]
    \ENDFOR
    \STATE Set
    $\hat{z}_{k+1}\leftarrow z_{k+1}^{(0)}$
\ENDFOR
\STATE \textbf{Output:}
$\hat{z}_1,\dots,\hat{z}_L$
\end{algorithmic}
\end{algorithm}

The training loss uses a fixed physiological coefficient
without additional timestep-dependent weighting.
The projection $\Pi_\phi$ consists of temporal resampling
and a learnable $1\times1$ convolution.
Default inference uses standard DDPM sampling
($K_{\mathrm{res}}=1$), without explicit energy-gradient
correction or energy-based resampling.
Physiological information is supplied through EPK
conditioning and the parameters learned under the
regularized training objective.

\section{Detailed Proofs for Energy-Regularized Diffusion}
\label{app:theory_proofs}

In this section, we provide the complete mathematical proofs for the theoretical claims. 

Let $p_\theta(z)$ denote the base empirical data distribution learned by the diffusion model, and $E(z)$ denote the parameterized energy function derived from the physiological ODE. Our target energy-regularized distribution is defined as:
\begin{equation}
    p^*(z) = \frac{1}{Z} p_\theta(z)\exp(-\gamma E(z)),
    \label{eq:target_dist}
\end{equation}
where $Z = \int p_\theta(z)\exp(-\gamma E(z)) dz$ is the intractable partition function, and $\gamma > 0$ is the temperature hyperparameter.

\subsection{Proof of the Variational Perspective (Gibbs Posterior)}
\label{app:proof_variational}

\paragraph{Proposition 1.} The target distribution $p^*(z)$ is the unique minimizer of the following variational free-energy objective:
\begin{equation}
    \mathcal{J}(q) = \mathrm{KL}(q \| p_\theta) + \gamma \mathbb{E}_q[E(z)].
\end{equation}

\paragraph{Proof.} 
We can expand the objective functional $\mathcal{J}(q)$ using the definitions of Kullback-Leibler (KL) divergence and expectation:
\begin{align}
    \mathcal{J}(q) &= \int q(z) \log \frac{q(z)}{p_\theta(z)} dz + \int q(z) \big( \gamma E(z) \big) dz \nonumber \\
    &= \int q(z) \left( \log q(z) - \log p_\theta(z) + \gamma E(z) \right) dz.
\end{align}
By adding and subtracting the log partition function $\log Z$, we can rewrite the terms inside the integral:
\begin{align}
    \mathcal{J}(q) &= \int q(z) \left( \log q(z) - \left( \log p_\theta(z) - \gamma E(z) - \log Z \right) - \log Z \right) dz.
\end{align}
Notice that from Equation~\ref{eq:target_dist}, we have $\log p^*(z) = \log p_\theta(z) - \gamma E(z) - \log Z$. Substituting this back into the equation yields:
\begin{align}
    \mathcal{J}(q) &= \int q(z) \left( \log q(z) - \log p^*(z) \right) dz - \log Z \int q(z) dz \nonumber \\
    &= \mathrm{KL}(q \| p^*) - \log Z.
\end{align}
Since $\log Z$ is a constant independent of the distribution $q$, minimizing $\mathcal{J}(q)$ is strictly equivalent to minimizing $\mathrm{KL}(q \| p^*)$. According to Gibbs' inequality, the KL divergence is non-negative ($\mathrm{KL} \ge 0$) and achieves its unique global minimum of zero if and only if $q(z) = p^*(z)$ almost everywhere. 
Therefore, $p^* = \arg\min_q \mathcal{J}(q)$. \hfill $\blacksquare$

\subsection{Proof of Score Function Equivalence}
\label{app:proof_score}

\paragraph{Proposition 2.} The score function of the energy-regularized distribution $p^*(z)$ is given by:
\begin{equation}
    \nabla_z \log p^*(z) = \nabla_z \log p_\theta(z) - \gamma \nabla_z E(z).
\end{equation}

\paragraph{Proof.}
Starting from the definition of the target distribution in Equation~\ref{eq:target_dist}, we take the natural logarithm of both sides:
\begin{equation}
    \log p^*(z) = \log p_\theta(z) - \gamma E(z) - \log Z.
\end{equation}
Next, we apply the gradient operator $\nabla_z$ with respect to the latent variable $z$. Since the partition function $Z$ is a constant scalar with respect to $z$, its gradient is zero ($\nabla_z \log Z = 0$). Thus, we immediately obtain:
\begin{equation}
    \nabla_z \log p^*(z) = \nabla_z \log p_\theta(z) - \gamma \nabla_z E(z).
\end{equation}
This confirms that incorporating the ODE energy constraint during generation requires only a straightforward additive modification to the score field learned by the base diffusion model. \hfill $\blacksquare$

\subsection{Proof of Stationary Distribution via the Fokker-Planck Equation}
\label{app:proof_fokker_planck}

\paragraph{Proposition 3.} Consider the modified Langevin sampling dynamics governed by the Stochastic Differential Equation (SDE):
\begin{equation}
    dz_t = \underbrace{\left( \nabla_z \log p_\theta(z_t) - \gamma \nabla_z E(z_t) \right)}_{\text{drift term } f(z_t)} dt + \sqrt{2}\, dW_t,
    \label{eq:langevin_sde}
\end{equation}
where $W_t$ is a standard Wiener process. \textit{(Note: Here, $t$ denotes the continuous integration time of the Langevin process, which is distinct from the MDP sequence step $k$ and the discrete diffusion step $\tau$.)} The target distribution $p^*(z)$ is a valid stationary distribution of this process.

\paragraph{Proof.} 
Let $p(z, t)$ denote the probability density of the state $z$ at time $t$. The time evolution of $p(z, t)$ under the SDE in Equation~\ref{eq:langevin_sde} is described by the Fokker-Planck equation (Kolmogorov forward equation):
\begin{equation}
    \frac{\partial p(z, t)}{\partial t} = -\nabla_z \cdot \big( f(z) p(z, t) \big) + \nabla_z \cdot \big( \nabla_z p(z, t) \big).
\end{equation}
We can express the right-hand side in terms of the probability flux $J(z, t)$:
\begin{equation}
    \frac{\partial p(z, t)}{\partial t} = -\nabla_z \cdot J(z, t), \quad \text{where} \quad J(z, t) = f(z) p(z, t) - \nabla_z p(z, t).
\end{equation}
A distribution $p(z)$ is a stationary distribution if the probability density does not change over time, i.e., $\frac{\partial p(z)}{\partial t} = 0$. A sufficient condition for this is that the probability flux vanishes everywhere: $J(z) = 0$.

We now evaluate the probability flux at our target distribution $p^*(z)$. Using the result from Proposition 2, we can rewrite the drift term $f(z)$ as:
\begin{equation}
    f(z) = \nabla_z \log p_\theta(z) - \gamma \nabla_z E(z) = \nabla_z \log p^*(z).
\end{equation}
Substituting $p^*(z)$ and $f(z) = \nabla_z \log p^*(z)$ into the expression for the probability flux:
\begin{align}
    J^*(z) &= f(z) p^*(z) - \nabla_z p^*(z) \nonumber \\
           &= \big(\nabla_z \log p^*(z)\big) p^*(z) - \nabla_z p^*(z).
\end{align}
By applying the standard calculus identity for the derivative of a logarithm, $\nabla_z \log p^*(z) = \frac{\nabla_z p^*(z)}{p^*(z)}$, we have $\big(\nabla_z \log p^*(z)\big) p^*(z) = \nabla_z p^*(z)$. Therefore:
\begin{equation}
    J^*(z) = \nabla_z p^*(z) - \nabla_z p^*(z) = 0.
\end{equation}
Since the probability flux $J^*(z)$ is identically zero for all $z$, the divergence $\nabla_z \cdot J^*(z)$ is also zero. Consequently, $\frac{\partial p^*}{\partial t} = 0$. 

This establishes $p^*(z)$ as a valid stationary target distribution for the modified Langevin dynamics. Note that rigorously proving unique asymptotic global convergence from an arbitrary initialization requires standard additional structural assumptions on the energy landscape (e.g., ergodicity, log-Sobolev inequalities, or strict positivity), which we assume hold practically within the continuous data support. \hfill $\blacksquare$

\section{Additional Evaluation of Physiological Fidelity and Predictive Consistency}
\label{app:additional_validation}

We complement the main experiments with analyses of pharmacological
mapping, twelve-lead interval fidelity, diagnostic risk changes,
autoregressive rollout, and inference-time energy sampling.
These evaluations distinguish the effect of the drug-to-ODE mapping
from removal of the EPK architecture, assess interval values beyond
normality agreement, and examine whether generated trajectories
preserve observed diagnostic changes and remain stable during
recursive prediction.

\subsection{Millisecond-Level Interval Fidelity against Official DiffuSETS}
\label{app:interval_value_fidelity}

To assess numerical fidelity beyond normality categories, we
compare the full EPK world model with the official DiffuSETS
\texttt{unet\_all} checkpoint on the same 1,822 evaluation pairs.
For an observed interval $u$ and generated interval $\hat{u}$,
value accuracy requires
\begin{equation}
|\hat{u}-u|
\leq
\max\!\left(25\,\mathrm{ms},\,0.05|u|\right).
\label{eq:interval_value_accuracy}
\end{equation}
The DiffuSETS intervals are re-scored from saved outputs without
regenerating the ECGs.
The full EPK results aggregate three training seeds: 17, 42, and 2026.

% Source: runs/table2_epk_module_ablation_20260923/
% aligned_vs_diffusets/aligned_comparison.md
% The meaning of +/- must follow the original run records;
% the supplied summary does not identify SD versus SE.
\begin{table}[htbp]
\centering
\caption{
Interval-value accuracy (\%) and signal MSE on 1,822 evaluation
pairs. Value accuracy follows
Equation~\ref{eq:interval_value_accuracy} and is distinct from
normality accuracy.
Full EPK results summarize three training seeds.
}
\label{tab:aligned_interval_value_fidelity}
\small
\setlength{\tabcolsep}{5pt}
\begin{tabular}{lrrrr}
\toprule
Method
& QTc $\uparrow$ & PR $\uparrow$ & TpTe $\uparrow$
& Signal MSE $\downarrow$ \\
\midrule
DiffuSETS official
& 17.7 & 44.2 & 66.5 & 0.087 \\
Ours, full EPK
& $\mathbf{55.8\pm0.2}$
& $\mathbf{78.4\pm0.7}$
& $\mathbf{93.4\pm0.3}$
& $\mathbf{0.057\pm0.003}$ \\
\bottomrule
\end{tabular}
\end{table}

As shown in Table~\ref{tab:aligned_interval_value_fidelity},
Ours improves value accuracy over the official DiffuSETS
checkpoint by 38.1, 34.2, and 26.9 percentage points for
QTc, PR, and TpTe, respectively, while reducing signal MSE
from 0.087 to 0.057.
These gains demonstrate stronger numerical interval fidelity
and waveform reconstruction for the complete EPK world model
under this evaluation, beyond agreement on normality categories.

The distinction between the two accuracy definitions is material.
On this split, which contains no reference QTc abnormalities,
the same saved DiffuSETS outputs yield 88.53\% QTc normality
accuracy but 17.7\% value accuracy.
Likewise, recall of normal intervals does not measure abnormality
sensitivity; PR balanced accuracy in the evaluated EPK runs is
approximately 50\%.
We therefore report normality agreement and interval-value accuracy
as different endpoints, without interpreting near-perfect normality
recall as evidence of reliable abnormality detection.

\subsection{EPK Architecture, Template Variation, and Cross-Lead Structure}
\label{app:epk_implementation_diagnostics}

\paragraph{Architecture-preserving controls.}
The EPK module uses 128-dimensional embeddings for cross-attention
conditioning and a learned energy projection
implemented as a $1\times1$ convolution from one input channel
to four output channels, with weight shape $[4,1,1]$.
Writing the pre-projection ODE sequence as $e_{\mathrm{EPK}}$,
the projected anchor is
$z_{\mathrm{EPK}}=\Pi_{\phi}(e_{\mathrm{EPK}})$.
Our architecture-preserving controls retain both EPK pathways and
vary the drug-to-ODE mapping or the energy regularization weight
$\lambda_{\mathrm{EPK}}$.
Disabling \texttt{use\_epk} removes the projection and is therefore
not equivalent to changing the mapping or setting only the energy
weight to zero.

\paragraph{Regimen-dependent template variation.}
Across eight drug regimens, the pharmacological mapping produces
nonzero variation in the pre-projection ODE templates, whereas the
identity mapping produces the same template for every regimen
(Table~\ref{tab:epk_template_variation}).
The permutation control preserves the template set while changing
its assignment to regimens, and therefore retains the same aggregate
variance and pairwise distance as the pharmacological mapping.
These diagnostics establish that the pharmacological map supplies
regimen-dependent templates rather than a repeated identity template.
We evaluate the pharmacological--identity difference separately
through generated intervals; the permutation statistics here
serve only as a template-variation control.

% Source: runs/rebuttal_zepk_anchor_stats.json
\begin{table}[htbp]
\centering
\caption{
Pre-projection EPK template variation across eight drug regimens.
Permutation preserves aggregate template variation by construction;
these statistics characterize template diversity, not the correctness
of drug--template assignments.
}
\label{tab:epk_template_variation}
\small
\begin{tabular}{lrr}
\toprule
Mapping & Mean variance & Mean pairwise $\ell_2$ distance \\
\midrule
Pharmacological & 0.004979 & 1.091 \\
Identity        & 0        & 0     \\
Permuted        & 0.004979 & 1.091 \\
\bottomrule
\end{tabular}
\end{table}

\paragraph{Projected anchors and multi-lead generation.}
The projected anchor has a mean cross-drug variance of
$1.66\times10^{-8}$, indicating low cross-drug variability on
the recorded scale; the separate EPK conditioning stream remains
active. We therefore assess the practical benefit at the generated
ECG level rather than inferring it from template diversity alone.

The one-dimensional McSharry prior supplies temporal structure,
while the world-model UNet and VAE decoder produce lead-specific
morphology.
The complete world model achieves a cross-lead correlation-matrix
Frobenius error of 2.69 relative to the observed ECG, compared with
6.39 when decoding the projected ODE template alone.
This corresponds to a 57.9\% lower error and shows that the learned
generator captures cross-lead relationships more accurately than
template-only decoding.
This supports the role of the learned generator in complementing
the temporal prior with multi-lead structure.

\subsection{Twelve-Lead Interval Effects of Pharmacological Mapping}
\label{app:epk_mapping_diagnostic}

We compare the pharmacological and identity mappings under the
\texttt{dadm\_table1} evaluation protocol, using the respective
\texttt{unet\_best.pth} checkpoints from the
\texttt{map\_pharm\_1} and \texttt{map\_identity\_1} runs.
Both variants retain the EPK conditioning stream and energy
projection.
Accuracy in this diagnostic measures agreement between the
normal/abnormal categories of generated and observed intervals,
rather than millisecond matching.
Table~\ref{tab:epk_mapping_all_leads} reports all 12 leads and
three interval metrics, with
$\Delta=\mathrm{Acc}_{\mathrm{pharm}}-\mathrm{Acc}_{\mathrm{id}}$
expressed in percentage points.

% Source: runs/rebuttal_per_lead_pharm_vs_identity.json
\begin{table}[htbp]
\centering
\caption{
Twelve-lead interval normality accuracy (\%) for identity (Id.)
and pharmacological (Pharm.) EPK mappings.
Positive $\Delta$ values favor pharmacological mapping and are
highlighted in bold.
Differences are taken from the original evaluation output rather
than recomputed from the displayed rounded accuracies.
}
\label{tab:epk_mapping_all_leads}
\small
\setlength{\tabcolsep}{3pt}
\renewcommand{\arraystretch}{1.08}
\begin{tabular}{lrrrrrrrrr}
\toprule
& \multicolumn{3}{c}{QTc}
& \multicolumn{3}{c}{PR}
& \multicolumn{3}{c}{TpTe} \\
\cmidrule(lr){2-4}
\cmidrule(lr){5-7}
\cmidrule(lr){8-10}
Lead
& Id. & Pharm. & $\Delta$
& Id. & Pharm. & $\Delta$
& Id. & Pharm. & $\Delta$ \\
\midrule
I
& 99.69 & 99.63 & $-0.06$
& 97.08 & 96.02 & $-1.06$
& 98.50 & 98.19 & $-0.31$ \\
II
& 99.69 & 99.63 & $-0.06$
& 95.59 & 92.73 & $-2.86$
& 98.94 & 98.07 & $-0.87$ \\
III
& 98.70 & 99.19 & $\mathbf{+0.50}$
& 76.78 & 77.75 & $\mathbf{+0.97}$
& 99.42 & 99.30 & $-0.13$ \\
aVR
& 98.27 & 98.03 & $-0.24$
& 52.05 & 52.27 & $\mathbf{+0.22}$
& 81.56 & 82.27 & $\mathbf{+0.71}$ \\
aVL
& 99.19 & 99.13 & $-0.06$
& 94.79 & 93.48 & $-1.30$
& 97.08 & 96.83 & $-0.25$ \\
aVF
& 99.44 & 99.25 & $-0.19$
& 75.79 & 83.36 & $\mathbf{+7.57}$
& 98.94 & 98.62 & $-0.31$ \\
V1
& 98.20 & 98.20 & $0.00$
& 72.86 & 69.42 & $-3.44$
& 55.29 & 56.45 & $\mathbf{+1.16}$ \\
V2
& 97.95 & 97.95 & $0.00$
& 70.02 & 70.40 & $\mathbf{+0.38}$
& 79.59 & 82.17 & $\mathbf{+2.58}$ \\
V3
& 98.13 & 98.07 & $-0.07$
& 98.20 & 98.57 & $\mathbf{+0.37}$
& 82.02 & 85.21 & $\mathbf{+3.19}$ \\
V4
& 99.50 & 99.31 & $-0.19$
& 95.65 & 94.64 & $-1.01$
& 92.48 & 88.65 & $-3.83$ \\
V5
& 99.32 & 99.13 & $-0.19$
& 95.03 & 93.85 & $-1.18$
& 97.33 & 95.21 & $-2.11$ \\
V6
& 99.69 & 99.50 & $-0.19$
& 93.54 & 91.80 & $-1.74$
& 99.07 & 98.32 & $-0.75$ \\
\bottomrule
\end{tabular}
\end{table}

The largest improvement occurs for PR in lead aVF, where
accuracy increases from 75.79\% to 83.36\%
(+7.57 percentage points).
Additional gains appear for QTc and PR in lead III, TpTe in
aVR and V1, and PR and TpTe in V2/V3.
In particular, TpTe accuracy improves by 2.58 and 3.19 percentage
points in V2 and V3, respectively.
The mapping effects are heterogeneous: TpTe accuracy decreases
by 3.83 and 2.11 percentage points in V4 and V5, and several
other endpoints favor identity.
Thus, this comparison identifies specific lead--interval benefits
from pharmacological modulation rather than uniform twelve-lead
superiority.
QTc accuracy near 99\% on most leads reflects normality agreement
under a low abnormal rate; numerical fidelity is assessed
separately using Equation~\ref{eq:interval_value_accuracy}.

\subsection{Agreement with Observed Post-Dose Diagnostic Changes}
\label{e4}

This evaluation examines whether generated ECGs preserve
the magnitude and direction of diagnostic-score changes
observed after medication administration, complementing
the waveform and interval evaluations.

\paragraph{Evaluation protocol.}
We use the frozen \texttt{ecg\_fm\_diagnostic} backend
described in Appendix~B.10 to score both generated and
observed ECGs. Given the predicted probabilities of
17 diagnostic labels, we define
\begin{equation}
R(x)=\frac{1}{17}\sum_{j=1}^{17}p_j(x).
\label{eq:diagnostic_mean_score}
\end{equation}
For each case $i$ and its recorded medication
$a_i^{\mathrm{true}}$, we generate $S=5$ standard-DDPM
samples with EPK conditioning and compute
\begin{equation}
\mu_{R,i}\!\left(a_i^{\mathrm{true}}\right)
=
\frac{1}{S}\sum_{s=1}^{S}
R\!\left(
f_{\mathrm{dec}}\!\left(
\hat z_i^{(s)}(a_i^{\mathrm{true}})
\right)
\right).
\label{eq:diagnostic_sample_mean}
\end{equation}
Predicted and observed diagnostic changes are measured
relative to the same pre-dose ECG:
\begin{align}
\Delta_{\mathrm{wm},i}
&=
\mu_{R,i}\!\left(a_i^{\mathrm{true}}\right)
-
R(x_{\mathrm{pre},i}),
\\
\Delta_{\mathrm{real},i}
&=
R(x_{\mathrm{post},i})
-
R(x_{\mathrm{pre},i}).
\label{eq:observed_diagnostic_change}
\end{align}
We assess their agreement using Pearson correlation,
Spearman correlation, and sign agreement.
The observed change is computed from the recorded
post-dose ECG, providing a reference independent of
the generated trajectories.

\paragraph{Consistency across repeated stochastic sampling.}
We evaluate the same \texttt{unet\_best.pth} checkpoint
from \texttt{ewm\_world\_8} on the same 200 test cases
across six sampling runs.
Each run independently regenerates its DDPM samples.
Although the evaluation files record different
action-selection coefficients $\lambda$, this coefficient
does not enter the diagnostic-change calculation for
a fixed recorded medication.
The results therefore characterize sensitivity to
stochastic sampling rather than the effect of $\lambda$.

As shown in Table~\ref{tab:diagnostic_change_runs},
all six runs exhibit positive agreement with observed
diagnostic changes: Pearson correlation ranges from
0.562 to 0.620, Spearman correlation from 0.545 to 0.598,
and sign agreement from 69.5\% to 75.5\%.
Across runs, the corresponding means are
$0.592\pm0.021$, $0.577\pm0.022$, and
$(71.3\pm2.2)\%$, respectively.
The relatively small between-run variation indicates
that the observed diagnostic-change agreement is
maintained across repeated stochastic generation,
rather than being confined to a single sampling run.

\begin{table}[t]
\centering
\small
\caption{
Agreement between simulated and observed diagnostic-score
changes on the same 200 test cases across six stochastic
sampling runs. Each run regenerates five DDPM samples
per recorded medication. Recorded $\lambda$ values apply
only to action selection and do not enter these metrics.
The final row reports the mean and sample standard
deviation across runs.
}
\label{tab:diagnostic_change_runs}
\begin{tabular}{ccccc}
\toprule
Run
& Recorded $\lambda$
& Pearson $r$
& Spearman $\rho$
& Sign agreement (\%) \\
\midrule
1 & 0.0 & 0.587 & 0.585 & 71.5 \\
2 & 0.2 & 0.613 & 0.597 & 71.0 \\
3 & 0.4 & 0.581 & 0.557 & 70.0 \\
4 & 0.6 & 0.620 & 0.598 & 75.5 \\
5 & 0.8 & 0.562 & 0.545 & 69.5 \\
6 & 1.0 & 0.591 & 0.580 & 70.5 \\
\midrule
Mean $\pm$ SD
& --
& $0.592\pm0.021$
& $0.577\pm0.022$
& $71.3\pm2.2$ \\
\bottomrule
\end{tabular}
\end{table}

\paragraph{Relationship to the main comparison.}
Table~\ref{tab:model_comparison} reports the correlation
and sign agreement from Run~4, in which our model
outperforms DADM on both correlation metrics and
directional agreement.
The six-run results provide additional evidence that
our model consistently preserves diagnostic-change
information under repeated sampling.
These standard deviations quantify sampling-related
variation for a fixed checkpoint and test cohort;
they do not estimate training-seed variability or
population-level confidence intervals.

\paragraph{Interpretation.}
Together with the waveform and interval results,
these findings support the ability of the ECG World
Model to preserve diagnostically relevant changes
under recorded medication conditions.
Because predicted and observed changes share the
pre-dose reference and use the same frozen diagnostic
backend, their correlations should be interpreted
as diagnostic-change fidelity within this evaluation
protocol.
They do not establish calibrated clinical risk,
causal treatment effects, or the benefit of selecting
an alternative medication.

\subsection{Short-Chain Autoregressive Rollout Stability}
\label{e5}

For each subject--admission pair, identified by \texttt{subject\_id}
and \texttt{hadm\_id}, we retain the longest recorded chain.
Autoregressive execution recursively generates the preceding states.
The oracle instead computes the final prediction using the true
preceding latent.
Let $\overline{E}_{\mathrm{roll}}$ and
$\overline{E}_{\mathrm{oracle}}$ be the corresponding mean
final-step latent MSEs.
We report
\begin{equation}
\begin{aligned}
\Delta_{\mathrm{roll}}
&=\overline{E}_{\mathrm{roll}}
  -\overline{E}_{\mathrm{oracle}},\\
\mathrm{Retention}
&=100\,
  \frac{\overline{E}_{\mathrm{oracle}}}
       {\overline{E}_{\mathrm{roll}}}.
\end{aligned}
\label{eq:rollout_retention}
\end{equation}
Retention is a ratio of mean errors, not a classification
accuracy or a mean of per-chain error ratios.

% Sources: results/eval_wm_rollout_retention_path_eq_ewm8_test_long.json
% results/eval_wm_rollout_retention_path_eq_ewm4_test_long_true.json
% results/eval_wm_rollout_retention_path_eq_ewm8_test_long100.json
% results/eval_wm_rollout_retention_path_eq_qwen_only100.json
\begin{table}[htbp]
\centering
\caption{
Final-step autoregressive rollout relative to oracle prediction.
The 1,273-chain evaluations use \texttt{unet\_best}; the separate
100-chain Ours evaluation uses \texttt{unet\_500}.
Qwen2.5-VL is an auxiliary untrained latent-prediction control,
with parsing limitations discussed in the text.
}
\label{tab:rollout_retention_diagnostic}
\small
\setlength{\tabcolsep}{4pt}
\begin{tabular}{llrrr}
\toprule
Model & Checkpoint & Chains
& $\Delta_{\mathrm{roll}}$ & Retention (\%) \\
\midrule
Ours (\texttt{ewm8}) & \texttt{unet\_best}
& 1,273 & $+0.000882$ & 98.7 \\
Ours (\texttt{ewm4}) & \texttt{unet\_best}
& 1,273 & $-0.000171$ & 100.3 \\
\midrule
Ours (\texttt{ewm8}) & \texttt{unet\_500}
& 100 & $-0.000030$ & 100.0 \\
Qwen2.5-VL-7B & ---
& 100 & $+0.002576$ & 93.8 \\
\bottomrule
\end{tabular}
\end{table}

On 1,273 hospitalization chains, \texttt{ewm8} achieves 98.7\%
retention with an additional mean latent MSE of only 0.000882.
The \texttt{ewm4} checkpoint has a slightly negative error
difference of $-0.000171$ and 100.3\% retention.
Both results indicate a small average terminal-error difference
between recursive and oracle execution in this evaluation.
A negative difference is possible because the oracle supplies
the true preceding input to the same learned predictor, rather
than defining a mathematical lower bound on prediction error.

The evaluated horizons are short: for the 1,273-chain
\texttt{ewm8} run, chain length in rows has median 2, mean 2.16,
and maximum 6.
Approximately 53\% of chains have positive error differences,
while the mean remains below 0.001 latent MSE.
Accordingly, these results demonstrate limited additional
terminal error over the observed hospitalization chains,
not an absence of absolute prediction error or a guarantee
for substantially longer horizons.

\paragraph{Auxiliary language-model control.}
Under the separate 100-chain protocol, \texttt{ewm8} retains
100.0\% compared with 93.8\% for Qwen2.5-VL-7B.
However, Qwen is not trained on these ECG latents and incurs
296 parse-failure steps.
Its 1,273-chain run has 3,768 parse failures, 99.1\% retention,
and a median error difference of zero.
These failures limit interpretation of the language-model
comparison; the primary stability evidence is the Ours
rollout--oracle comparison, not superiority over a trained
ECG-specific dynamics baseline.

\subsection{Inference-Time Energy Sampling}
\label{e6}

Our default inference procedure uses standard DDPM sampling
with EPK conditioning from the model trained with physiological
energy regularization.
To examine an additional inference-time use of this prior,
we evaluate EPK energy resampling on 48 evaluation pairs,
using the same pharmacological checkpoint and 200 denoising
steps.
The resampling parameter $K_{\mathrm{res}}$ is distinct
from the sample counts used for reconstruction evaluation
and diagnostic-score estimation.

\begin{table}[t]
\centering
\caption{
Inference-time energy resampling on 48 evaluation pairs
using the same pharmacological checkpoint and 200 denoising
steps. Energy resampling is an optional inference procedure
with additional sampling cost.
}
\label{tab:inference_energy_pilot}
\small
\begin{tabular}{lrr}
\toprule
Sampler & Latent MSE $\downarrow$ & Signal MSE $\downarrow$ \\
\midrule
Standard DDPM ($K_{\mathrm{res}}=1$)
& 0.0543 & 0.0676 \\
EPK energy resampling ($K_{\mathrm{res}}=8$)
& \textbf{0.0492} & \textbf{0.0620} \\
\bottomrule
\end{tabular}
\end{table}

As shown in Table~\ref{tab:inference_energy_pilot},
EPK energy resampling reduces latent MSE by approximately
9.4\% and signal MSE by 8.3\% relative to standard DDPM.
These results demonstrate improved reconstruction under
the evaluated resampling configuration.
Because the comparison changes both the sampling procedure
and its computational budget, it does not isolate the
benefit of energy-based selection at matched cost.

A separate direct energy-gradient sweep on the same
48 pairs yields only a small change in latent MSE,
from 0.05023 to 0.05022 as the coefficient increases
from 0 to 0.1.
Thus, resampling provides the clearer empirical benefit
among the inference-time configurations tested here.
The energy-tilted distribution serves as a theoretical
reference; neither the training regularizer nor this
resampling pilot establishes exact sampling from that
distribution.
Standard DDPM remains the default inference procedure
for the reported main experiments.

\paragraph{Summary.}
The additional evaluations support complementary aspects
of physiology-informed ECG simulation:
improved interval-value accuracy relative to the official
DiffuSETS checkpoint, lead-specific benefits from
pharmacological mapping, improved cross-lead structure
relative to template-only decoding, agreement with observed
diagnostic-score changes, and limited additional terminal
error over the evaluated short hospitalization chains.
The resampling pilot further suggests that additional
inference-time computation can improve reconstruction.
Together, these findings support the model's predictive
fidelity across waveform, interval, diagnostic-change,
and short-chain prediction evaluations.

\subsection{Intervention Sets and Candidate Selection}
\label{app:action_sets}

\paragraph{Eight-regimen set for interval and EPK experiments.}
The controlled drug-response experiments use eight predefined
regimen labels implemented in
\texttt{dataset/ecg\_medication\_8class.py}:
Ranolazine, Dofetilide, Verapamil, Quinidine, Placebo,
Moxifloxacin+Diltiazem, Mexiletine+Dofetilide, and
Lidocaine+Dofetilide.
Labels 0--4 are drawn from \texttt{ecg\_data}, and labels
5--7 from \texttt{ecg\_data1}, corresponding to the
controlled drug-response cohorts described in Appendix~A.
Combination regimens are encoded as distinct categorical
actions.
This intervention vocabulary is predefined and is not
derived from MIMIC-IV medication frequencies.

\paragraph{Five-medication set for risk experiments.}
The risk experiments use a separate candidate set constructed
from \texttt{ewm\_world\_vae\_mortality.pt}.
The function
\texttt{compute\_topk\_test\_only\_drug\_texts}
counts medication names only among records with
\texttt{split=test}.
For each record, the name is taken from the label's
\texttt{text} field, falling back to \texttt{drug\_name}
when \texttt{text} is absent.
The procedure scans 21,328 test records containing 1,263
distinct medication-name strings and selects the five most
frequent names using \texttt{most\_common(5)}.
Training records do not contribute to this ranking.
Table~\ref{tab:risk_action_set} reports the resulting
candidate set.
Each name is mapped to the first matching
\texttt{medication\_id} found in the same data file.

\begin{table}[t]
\centering
\caption{
Fixed candidate medications for the MIMIC-IV risk experiments.
Counts denote occurrences in the full test split;
IDs are internal medication identifiers in the processed data.
}
\label{tab:risk_action_set}
\small
\begin{tabular}{lrr}
\hline
Medication & Test count & Medication ID \\
\hline
Propofol                      & 1,045 & 7  \\
Insulin - Regular             & 936   & 25 \\
Heparin Sodium (Prophylaxis)   & 848   & 22 \\
Furosemide (Lasix)             & 736   & 14 \\
Norepinephrine                & 726   & 19 \\
\hline
\end{tabular}
\end{table}

\paragraph{Evaluation scope.}
Candidate ranking is restricted to this fixed five-medication
set.
Top-1 concordance measures whether the selected candidate
matches the recorded medication, with a uniform random-choice
baseline of $1/5=0.20$ when the recorded medication belongs
to the candidate set.
The candidate vocabulary is determined from test-split
medication frequencies, rather than specified independently
of the test set.
No diagnostic outcomes are used for candidate selection.
Accordingly, this evaluation measures restricted-set
medication concordance and does not establish treatment
optimality or clinical safety.

\section{Impact Statements}

This work proposes an ECG World Model for simulation-based analysis of cardiac interventions, enabling AI systems to explore potential physiological responses under different treatment scenarios. By combining a physiological prior with data-driven generative modeling, the framework supports hypothesis generation and in-silico evaluation of treatment effects.

Several limitations should be noted. First, the model is trained on observational and simulated data, which may not fully capture real-world complexity, variability, or rare edge cases, potentially leading to deviations from true clinical outcomes under distributional shifts. Second, the physiological prior is simplified and may be misspecified, limiting its ability to represent full cardiac dynamics, especially in pathological regimes. Third, evaluation focuses on aggregate stability metrics rather than clinical endpoints, and does not fully explain the interaction between prior-driven and data-driven components. Finally, the model lacks calibrated uncertainty estimates and formal safety guarantees, constraining its use in high-stakes settings.

Accordingly, the system should not be used as a standalone clinical tool, but rather as a decision-support framework requiring expert validation. Generated scenarios should be treated as exploratory hypotheses rather than definitive predictions.

Despite these limitations, the model exhibits stable and bounded behavior across varying levels of prior mismatch, with variability increasing smoothly rather than collapsing. The influence of the prior is also continuous and controllable, as partial prior injection yields intermediate performance. These results suggest a degree of robustness under imperfect assumptions.

Overall, this work takes a step toward more robust and interpretable generative models for physiological simulation, while underscoring the need for further validation, mechanistic understanding, and safety-aware design prior to real-world deployment.

\section{Devices}
\label{devices}
In the experiments, we conducted all methods on a local Linux server equipped with an AMD EPYC 7742 64-Core Processor (128 logical threads). All methods are implemented using the PyTorch framework, and all models are trained on NVIDIA A800-SXM4-80G GPUs (80GB HBM2e memory).

\section{AI Use Statement}
In preparing this paper, we used large language models (LLMs) solely as an assistive tool for language polishing and writing improvements. The models were not involved in research ideation, experimental design, data analysis, or drawing scientific conclusions. The authors take full responsibility for the contents of this paper.
For experimental evaluation, we additionally conducted testing using large language models such as GPT.

\section{Reproducibility Statement}

We document the model formulation, physiological prior construction,
training objective, and inference procedure, together with dataset
sources, cohort construction, and subject-disjoint training and test
splits. The reported experimental settings include the main
optimization hyperparameters, physiological regularization settings,
stochastic sampling configurations, frozen pretrained components,
and computing environment. We also provide training and inference
pseudocode, mathematical derivations with their underlying
assumptions, and additional ablation and robustness evaluations.

Reproducing the experiments requires access to the corresponding
ECG and clinical datasets, pretrained components, and model services
used in the comparisons. Evaluations should preserve the reported
cohort definitions, subject-level split protocol, model configurations,
and sampling settings. Source data should be obtained and used in
accordance with the applicable access and usage conditions.

\section{Ethics Statement}

This work investigates intervention-aware cardiac simulation using
existing ECG and clinical datasets, including MIMIC-IV-ECG,
MIMIC-IV-Clinical, ECGRDVQ, and ECGDMMLD. The out-of-distribution
evaluation cohort is constructed from MIMIC-IV-ECG records.
The proposed framework is a proof-of-concept for hypothesis generation
and simulation-based analysis, rather than a standalone clinical tool
or a substitute for professional clinical judgment. Access to and
reuse of the source data should respect the corresponding data-use
conditions and patient-confidentiality requirements.

Generated ECG trajectories and intervention comparisons should be
interpreted as exploratory model predictions, not definitive evidence
of treatment efficacy or safety. The downstream risk scores are
proxies derived from ECG diagnostic predictions rather than directly
observed clinical outcomes. Similarly, variability across stochastic
simulations does not constitute a formal guarantee of clinical safety.
Simplified physiological priors, discrete intervention representations,
incomplete modeling of drug interactions, and distribution shifts may
limit the reliability of simulated responses. Broader clinical
validation and expert assessment are therefore required before any
real-world application, and model outputs should not be used
independently to prescribe medications or alter patient care.

\end{document}